%% file: main.tex
\documentclass[lettersize,journal]{IEEEtran}
\usepackage{amsmath,amsfonts}
\usepackage{algorithmic}
\usepackage{algorithm}
\usepackage{array}
\usepackage[caption=false,font=normalsize,labelfont=sf,textfont=sf]{subfig}
\usepackage{textcomp}
\usepackage{stfloats}
\usepackage{url}
\usepackage{verbatim}
\usepackage{graphicx}
\usepackage{cite}
\usepackage{bibunits}

\usepackage{tcolorbox,multirow}
\usepackage{xcolor,mdframed}
\usepackage{colortbl,booktabs}
\usepackage{pifont}
\usepackage{tablefootnote}
\usepackage{arydshln}
\usepackage{amssymb}
\usepackage{booktabs}
\usepackage{color} 

\usepackage{xspace}
\makeatletter
\newcommand{\thickhline}{%
    \noalign {\ifnum 0=`}\fi \hrule height 0.7pt
    \futurelet \reserved@a \@xhline
}

\newcolumntype{I}{!{\vrule width 0.8pt}}

\DeclareRobustCommand\onedot{\futurelet\@let@token\@onedot}
\def\@onedot{\ifx\@let@token.\else.\null\fi\xspace}

\newcolumntype{P}[1]{>{\centering\arraybackslash}p{#1}}
\definecolor{lightgray}{gray}{.9}
\definecolor{deepgray}{gray}{.8}
\definecolor{mygreen}{RGB}{29, 154, 120}
\definecolor{DarkGreen}{RGB}{42,110,63}

\defaultbibliographystyle{IEEEtran}
\defaultbibliography{main}

\begin{document}

\title{An Empirical Study of Validating Synthetic Data for Text-Based Person Retrieval}

\author{Min Cao, Yuxin Lu, Ziyin Zeng, Dong Yi, Jinqiao Wang, and Mang Ye
\thanks{This work is supported by the National Natural Science Foundation of China under Grants 62476188, the Natural Science Foundation of the Jiangsu Higher Education Institutions of China.}
\thanks{Min Cao, and Yuxin Lu are with the School of Computer Science and Technology, Soochow University, China (e-mail: caomin0719@126.com).}
\thanks{Ziyin Zeng is with Frontis.Al, Beiiing, China.}
\thanks{Dong Yi is with Institute of Automation, Chinese of Academy, and is the corresponding author.}
\thanks{Jinqiao Wang is with the Foundation Model Research Center, Institute of Automation, Chinese Academy of Sciences, also with the School of Artifcial Intelligence, University of Chinese Academy of Sciences, and also with the Wuhan AI Research, Wuhan, China.}
\thanks{Mang Ye is with the School of Computer Science, Wuhan University, China.}
}
\markboth{}{}


\maketitle

\begin{abstract}
Data plays a pivotal role in Text-Based Person Retrieval (TBPR) research.
Mainstream research paradigm necessitates real-world person images with manual textual annotations for training models, posing privacy concerns and annotation burdens.
Several pioneering efforts explore synthetic data generation, and yet still depend on real data as a foundation, inheriting the same limitations. 
The feasibility of purely synthetic TBPR data remains unexplored, and there is currently no systematic study on the effectiveness boundaries of synthetic data across various real-world scenarios.
In this work, we present the first comprehensive empirical study of synthetic data for TBPR, with two key aspects.
(1) We propose a unified data synthesis pipeline that can operate entirely without real person data.
It combines an inter-class image generation module that produces diverse identity-centric images by means of an automatic prompt construction strategy, and an intra-class augmentation module that enhances identity variation through text-driven image editing.
(2) Leveraging this pipeline and an automatic textual description generation, we explore the effectiveness of synthetic data in diverse scenarios through extensive experiments, to reveal its practical utility as either a standalone replacement or a complementary augmentation to real data.
We release the code, along with the synthetic large-scale dataset generated by our pipeline in \url{https://github.com/Flame-Chasers/SynTBPR}.
\end{abstract}

\begin{IEEEkeywords}
Text-based person retrieval, data generation, empirical study.
\end{IEEEkeywords}

\begin{bibunit}
\input{sec/1_intro}
\input{sec/2_Related}

\input{sec/3_method}
\input{sec/4_experiment}

\input{sec/5_limitation}
\input{sec/6_conclusion}

\putbib
\end{bibunit}

\clearpage
\section*{Supplementary Material}

\begin{bibunit}
\input{sec/7_supplementary}
\putbib
\end{bibunit}

\end{document}

%% file: sec/1_intro.tex
\section{Introduction}
\label{sec:intro}
\begin{figure}
\setlength{\abovecaptionskip}{0.05cm}
\centerline{\includegraphics[width=1\linewidth,height=0.48\linewidth]{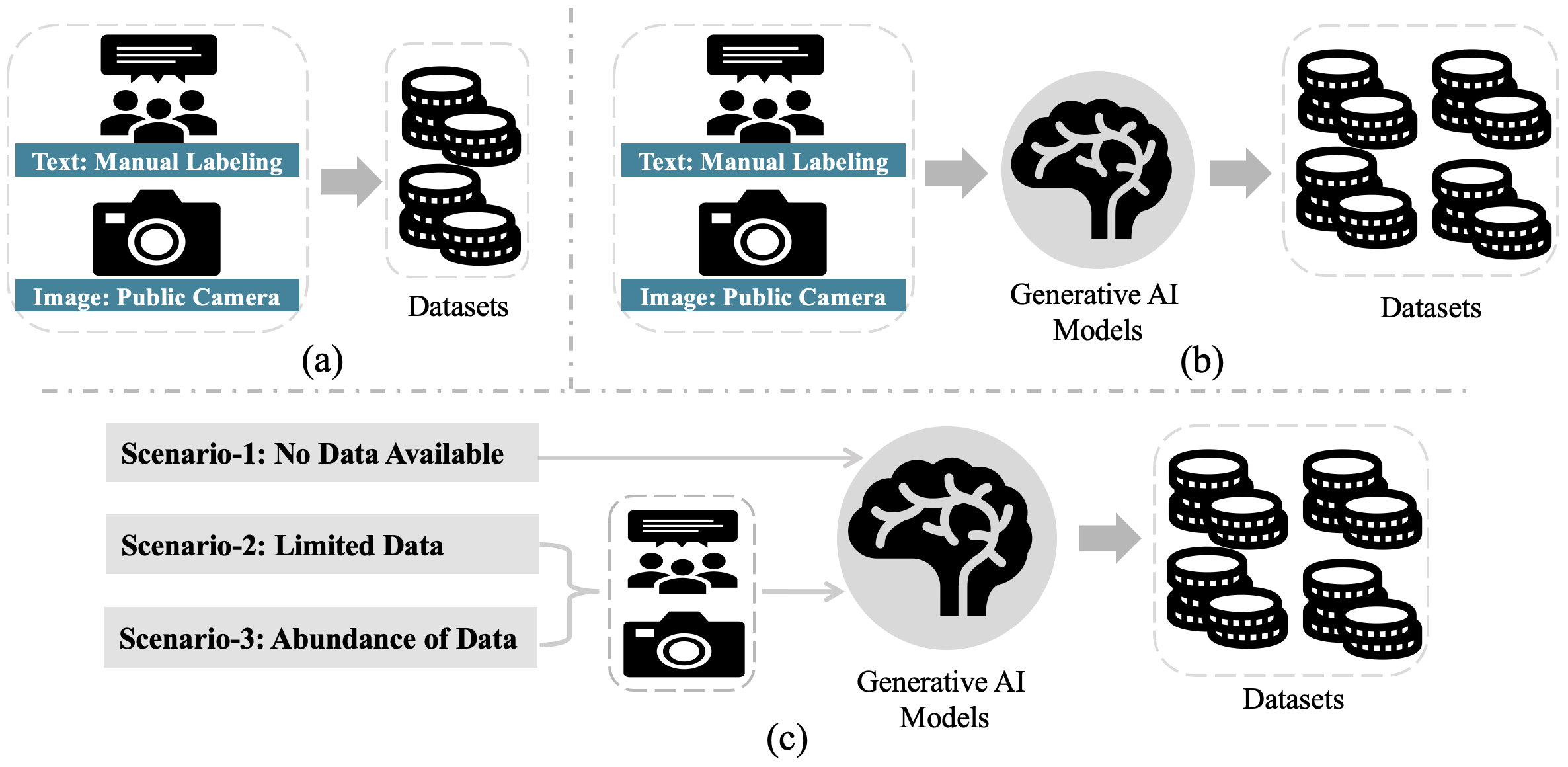}}
\caption{Data production paradigms for TBPR model training. 
(a) Data from a repository of real-world person images accompanied by manual textual annotations.
(b) Data produced by generative AI models with the assistance of real-world person images or manual textual annotations.
(c) Our proposed data production paradigm centered around three representative scenarios.}
\label{fig1}
\end{figure}

Text-Based Person Retrieval (TBPR)~\cite{li2017person} aims to identify a specific individual from a gallery of person images given a textual description.
With its wide range of potential applications, from missing person tracking to social media analysis, TBPR is experiencing rapid research progress~\cite{cao2024empirical,li2024adaptive,jiang2023cross,lu2025prompt,cao2025multilingual}.
Nevertheless, current research paradigm necessitates a substantial collection of real-world person images along with manual textual annotations to train models (Fig.~\ref{fig1} (a)), posing significant challenges related to privacy concern regarding person images and labor-intensive process required for textual annotations.

Recent advances in generative Artificial Intelligence (AI) models~\cite{yang2024qwen2,rombach2022high,bai2023qwen} have resulted in significant success in synthesizing high-quality data, revolutionizing various fields~\cite{he2022synthetic,dunlap2023using,tian2024stablerep}. This success raises the question: \emph{is synthetic data from generative AI models ready for TBPR task?}
There have been a few pioneering efforts~\cite{song2024diverse,wu2023contrastive,yang2023towards,tan2024harnessing,shao2023unified,zuo2024plip,jiang2025modeling} at exploring synthetic data from generative AI models to enhance TBPR performance (Fig.~\ref{fig1} (b)).
Wu~\emph{et al.}~\cite{wu2023contrastive} and Song~\emph{et al.}~\cite{song2024diverse} 
used generative AI to enhance real-data diversity for improved TBPR training.
Several works~\cite{tan2024harnessing,shao2023unified,zuo2024plip,jiang2025modeling} leveraged the large-scale real person dataset as image source to generate corresponding texts, thereby constructing pre-training TBPR dataset.
In contrast, Yang \emph{et al.}~\cite{yang2023towards} introduced a fully virtual pre-training dataset, with person images generated by an off-the-shelf diffusion model and a multimodal large language model from original textual annotations. 


While promising, these methods primarily generate synthetic data by relying on real data. Their effectiveness is typically demonstrated under data-rich assumptions, implicitly treating real data as a necessary prerequisite for synthesis.
Consequently, two critical gaps remain.
(1) The feasibility of purely synthetic TBPR data, \emph{i.e.,} data constructed without any access to real data, has not been systematically investigated, despite representing an extreme yet practically important scenario in privacy-constrained or annotation-scarce deployments.
(2) There is no empirical study that characterizes the effectiveness boundaries of synthetic data across varying levels of real-data availability, which is a crucial understanding for guiding the real-world application of TBPR under diverse data environments.

In this paper, we thoroughly and systematically study the effectiveness of synthetic data from generative AI models in TBPR, highlighting two main advancements.
1) We construct a unified data synthesis pipeline for TBPR that is fully functional even in the complete absence of real person data.
The pipeline consists of two components: an inter-class image generation module and an intra-class image augmentation module. The former synthesizes diverse identity-centric person images using text-to-image models guided by automatically constructed prompts; the latter enhances intra-identity variation through text-driven image editing.
2) We conduct a thorough empirical study on synthetic data for TBPR.
Leveraging the proposed data synthesis pipeline, along with an automatic text generation inspired by~\cite{tan2024harnessing}, we study the effectiveness of synthetic data in three representative scenarios:
(1) no data, aligning with privacy protection and labor liberation;
(2) limited data (\emph{e.g.,} $8$ pairs), for minimal-resource setting;
(3) abundant data, typical in academic settings.
Moreover, recognizing that synthetic data inevitably contain noise, we conduct an in-depth study employing various strategies to mitigate its effects and enhance performance.

In summary, rather than proposing a new TBPR method, we focus on an underexplored yet critical factor—data—and present two core contributions.
1) We present the first data synthesis pipeline for TBPR that supports scenario with no access to real person data.
2) We present the first systematic study of synthetic data in TBPR across the full spectrum of real-data availability. This study reveals the practical utility of synthetic data as either a standalone replacement or a complementary augmentation for real data.



%% file: sec/2_Related.tex
\section{Related work}
\label{sec:formatting}

\subsection{Text-based person retrieval}
TBPR involves retrieving a specific pedestrian from a vast image gallery using the text query. 
The task has evolved through advances in model architecture~\cite{bai2023rasa,jiang2023cross} and feature alignment~\cite{jiang2023cross,shao2022learning,zhang2018deep}.
Regarding model architecture, early methods employed VGG for image encoding~\cite{chen2018improving,li2017person} and LSTM for text encoding~\cite{zhang2018deep,yan2023image}, which later shifted to more robust encoders like ResNet-50 for image encoding~\cite{ding2021semantically,farooq2022axm,wang2020vitaa} and BERT for text encoding~\cite{li2022learning,zhu2021dssl}. Recently, methods based on CLIP~\cite{cao2024empirical,radford2021learning} and ALBEF~\cite{bai2023rasa,yang2023towards} have gained popularity for their joint image-text representation learning.
Regarding feature alignment, two types are usually explored: global alignment and local alignment. Early methods~\cite{shu2022see,wu2023refined} primarily focused on global alignment, aligning entire images with their paired textual descriptions. 
Recent methods~\cite{wang2022look,tang2022learning,qi2025granularity,you2025diverse} highlight local alignment. Jiang et al.~\cite{jiang2023cross} incorporated the masked language modeling to the proposed cross-modal implicit relation reasoning framework to establish local alignment.
In contrast to these TBPR methods, our goal is to explore the potential of synthetic data for TBPR in this work.

\subsection{Learning with synthetic data}
Exploring synthetic data to train machine learning models in various tasks is well studied~\cite{tian2024stablerep,fan2024scaling,he2022synthetic,dunlap2023using,huang2024unlabeled}.
In pedestrian-related tasks, early studies~\cite{wang2022cloning,barbosa2018looking,wang2020surpassing,zhang2021unrealperson,sun2019dissecting} used manual simulation of virtual environments and hand-crafted 3D person models from game engines to synthesize large-scale datasets. 
These methods have limitations, such as gaps with real-world data due to manual processes.
Subsequent studies~\cite{xiang2023less,zhong2018camstyle,wei2018person} focused on Generative Adversarial Networks (GANs)~\cite{goodfellow2014generative} to generate images. Thanks to their ability to model data distributions automatically, GANs partially mitigate the aforementioned limitations, yet often suffer from training instability.
Recently, diffusion-based models~\cite{song2024diverse,wu2023contrastive,yang2023towards,niu2024synthesizing,kim2024pose,yang2024qwen2} have gained attention for their superior training stability and ability to generate highly realistic images compared to GANs.

Specifically, the following works related to our study focus on synthesizing data for TBPR.
Wu~\emph{et al.}~\cite{wu2023contrastive} enhanced data diversity by generating new text–image pairs near the original samples in feature space.
Similarly, Song~\emph{et al.}~\cite{song2024diverse} edited original images using reference images of clothing and accessories.
Several works~\cite{tan2024harnessing,shao2023unified,zuo2024plip,jiang2025modeling} focused on generating synthetic textual descriptions from large-scale real person dataset~\cite{fu2021unsupervised,fu2022large,song2018region} to construct pre-training TBPR dataset.
In comparison, Yang~\emph{et al.}~\cite{yang2023towards} suggested constructing the pre-training dataset by directly generating person images and texts, but requiring original texts to guide diffusion-based models.
These works rely on original data to generate new data, leading to labor-intensive and privacy-sensitive issues.
In addition, they explore in a one-sided manner, lacking a comprehensive exploration of synthetic data in TBPR.

\begin{figure*}
\setlength{\abovecaptionskip}{0.05cm}
\centerline{\includegraphics[width=1\linewidth,height=0.4\linewidth]{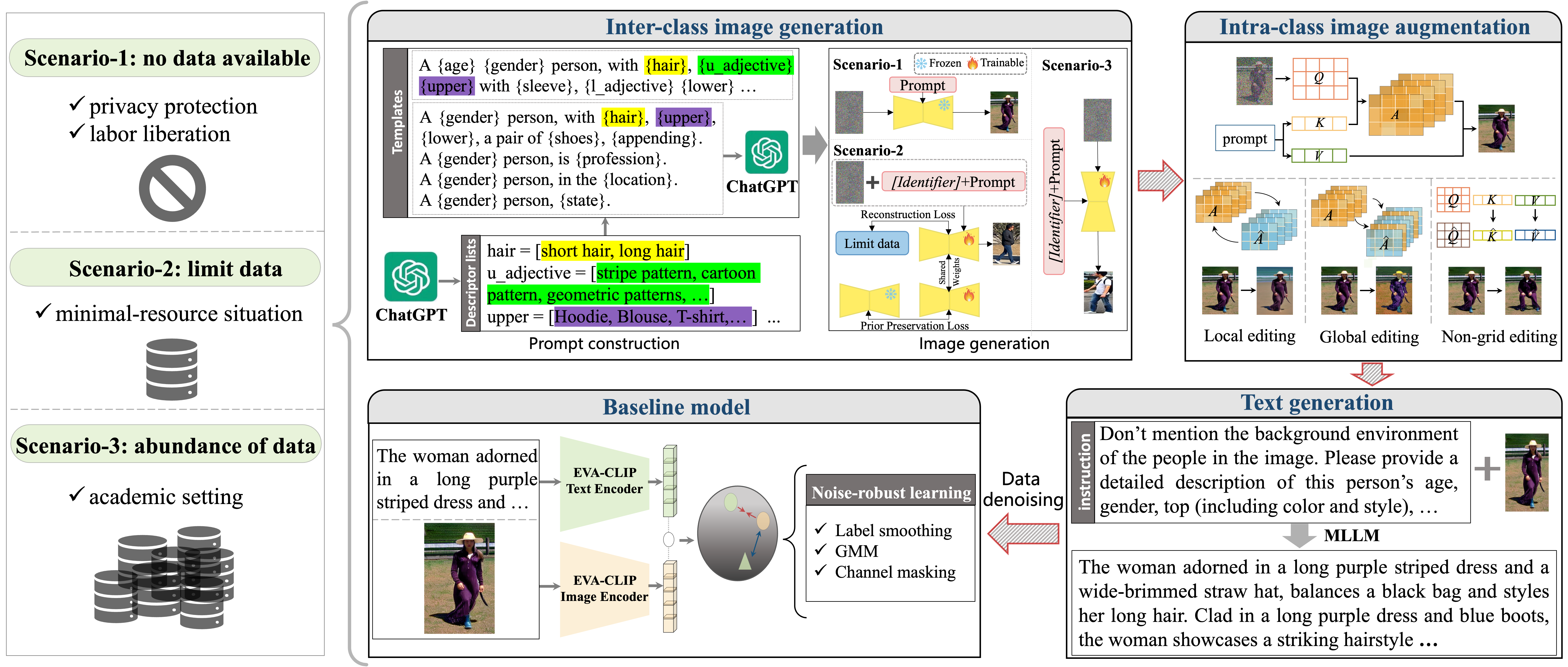}}
\caption{Workflow of our framework for validating synthetic data for TBPR.
It involve the following steps.
(1) Inter-class image generation: producing diverse person images with different identities;
(2) Intra-class image augmentation: further multiple variants per identity;
(3) Text generation: extracting textual descriptions of the person images;
(4) Baseline model: training the model using synthetic data alongside original real data (if accessible).
The framework is performed across three representative scenarios.}
\label{fig2}
\end{figure*}

%% file: sec/3_method.tex
\section{Methodology}

Most researchers currently  develop and train TBPR methods using public benchmarks from real surveillance cameras with manual annotations, denoted as $D^{Real}_{train} = \{(I_i, T_i)\}_{i=1}^{N}$.
This work focuses on generating data, $D^{Syn}_{train}$. 
We examine three generation scenarios:
(1) no $D^{Real}_{train}$ (\emph{i.e.,} $N = 0$), 
(2) a very small-scale $D^{Real}_{train}$ (\emph{e.g.,} $N = 8$), 
(3) unaltered $D^{Real}_{train}$. 
The TBPR model is then trained on $D_{train} = D^{Real}_{train} + D^{Syn}_{train}$.
Fig.~\ref{fig2} illustrates our overall framework.

\subsection{Inter-class image generation}
As the first component of our unified data synthesis pipeline, the inter-class image generation module aims to produce diverse person images with different identities.

\textbf{Prompt construction.}
The generative AI models are employed for image generation, with effective prompt design as a key technical aspect.
Unlike prior methods that use textual annotations from TBPR datasets as prompts~\cite{song2024diverse}, we propose an automatic prompt construction strategy, eliminating the need for labor-intensive annotation.

To ensure diverse person image generation, it is crucial to have plentiful prompts that cover detailed appearance descriptions of varied individuals. 
To this end, we design an intuitive plain description template that outlines the pedestrian from head to toe, as follows:

\begin{tcolorbox}[colback=white, colframe=black, opacityframe=0.5]
\small \emph{A \{age\} \{gender\} person, with \{hair\}, \{u\_adjective\} \{upper\} with \{sleeve\}, \{l\_adjective\} \{lower\}, a pair of \{shoes\}, \{appending\}, \{angle\}.}
\end{tcolorbox}

The various prompts are obtained from the template,
by randomly replacing each $\{*\}$  with an element from predefined descriptor lists.
These lists, created by ChatGPT, provide common and explicit descriptions of individuals
(\emph{e.g.,} \emph{\{lower\}=[trousers, skirt, dress, pants, jeans, shorts, leggings]}).
More details are in the supplementary material.

Nevertheless, the diversity of synthetic images can reach a bottleneck as generation scales up using the template. 
This is expected: a fixed structural template with a finite descriptor set provides only limited guidance for image synthesis.
For this, we further develop the Large Language Model (LLM)-extended prompts as a supplement.
Specifically, we design several rough description templates from different aspects, including appearance, profession, location and state, as follows:
\begin{tcolorbox}[colback=white, colframe=black, opacityframe=0.5]
\begin{itemize}
    \item \small \emph{A \{gender\} person, with \{hair\}, \{upper\}, \{lower\}, a pair of \{shoes\}, \{appending\}.}
    \item \small \emph{A \{gender\} person, is \{profession\}.}
    \item \small \emph{A \{gender\} person, in the \{location\}.}
    \item \small \emph{A \{gender\} person, \{state\}.}
\end{itemize}
\end{tcolorbox}

Following the same process, we first generate prompts by replacing each $\{*\}$ in the templates with an element from the descriptor lists.
We then instruct Qwen2~\cite{yang2024qwen2} to extend these primary prompts into final, more diverse ones.
Details on constructing LLM-extended prompts are provided in the supplementary material.


In conclusion, prompts are generated from five templates: one plain and four rough. The plain template directly describes pedestrian appearance, while rough templates, extendable by LLM, offer greater variety. 
To better convey the richness of our prompt design, Table~\ref{tab-prompt} summarizes the number of descriptor terms in each category, offering an intuitive sense of the expressive variety in appearance descriptions.
Unlike methods that rely on extensive manual annotations, our approach leverages a small set of designed templates, significantly reducing human effort.


\textbf{Image generation.}
These prompts guide person image generation using a generative AI model. Three scenarios are considered: no real person data, limited data access, and full TBPR dataset access. 
Specifically, we employ Stable Diffusion (SD)~\cite{rombach2022high} as the baseline, adapted for each scenario.

In the first scenario, without real training data, images are generated by applying constructed prompts to the offline SD. Varying prompts and random seeds yield a large, diverse set of synthetic person images. 
In the last scenario, SD is first fine-tuned with LoRA~\cite{hu2021lora} on all available data, then generates images from constructed prompts.
A technical token \emph{[identifier]} is added to the text prompt in fine-tuning.
The token acts as a unique identifier for the subject of person, enabling SD to generate new appearances of the person in inference.
Thus, \emph{[identifier]} must have a weak prior in training SD, for which is tokenized from rare vocabulary tokens.

In the second scenario, with limited training data available, we train SD with the following loss referring to~\cite{ruiz2023dreambooth},
\begin{equation}
\begin{aligned}
     \mathbb{E}_{{I}, c, \varepsilon, \varepsilon', t} &[ \omega_t \left\| \hat{{I}}_{\theta}(\alpha_t {I} + \sigma_t \varepsilon, c) - {I}\right\|^2_2 \ + \\
     & \qquad \lambda \omega_t' \left\| \hat{{I}}_{\theta}(\alpha_t' {I}_{pr} + \sigma_t' \varepsilon', c_{pr}) - {I}_{pr}\right\|^2_2].
    \label{twoS-loss}
\end{aligned}
\end{equation}

\begin{table*}
\setlength{\abovecaptionskip}{0.1cm}
    \centering
    \caption{The number of each descriptor term in the prompt.}
    \scalebox{0.95}{
    \begin{tabular}{cccccccccccccc}
            \toprule
            \rowcolor{lightgray}
            age	& gender & hair & u\_adjective & upper & sleeve & l\_adjective & lower & shoes	& appending & angle & profession & location & state \\
            \midrule
            3 & 2 & 2 & 5 & 11 & 2 & 38 & 7 & 5 & 6 & 5 & 64 & 96 & 258 \\
            \bottomrule
        \end{tabular}
    }
    \label{tab-prompt}
\end{table*}

The first loss term is referred to as the reconstruction loss.
$I$ is the ground-truth image and $\hat{{I}}_{\theta}({I}_t,c)$ with ${I}_t=\alpha_t {I} + \sigma_t \varepsilon$ is the predicted image conditioned on $c=\mathcal{F}(p)$ at $t$ time.
$c$ is the feature representation of text prompt $p$ encoded by a text encoder $\mathcal{F}$.  
$\varepsilon \sim \mathcal{N}([0,1])$ is the noise map. 
$\omega_t$, $\alpha_t$ and $\sigma_t$ are functions of the diffusion process time $t \sim \mu([0,1])$ and impact the quality of the samples.
A specific design is to add the token \emph{[identifier]} before the word \emph{`person'} in text prompt $p$.
The second loss term is called the class-specific prior preservation loss. $I_{pr}$ is the generated images conditioned on $c_{pr}$ by the off-line SD model and $c_{pr}$ is the representation of original text prompt (without the token \emph{[identifier]}).  
$\lambda$ is the hyper-parameter of weight.
This loss aims to counteract overfitting in SD during few-shot fine-tuning, helping retain the model's original language understanding and generation diversity.
A visualization of Eq. (1) is provided in the supplementary material to aid understanding.

\begin{table*}
\setlength{\abovecaptionskip}{0.15cm}
    \centering
    \caption{Prompts used in image editing. The original prompt used for editing the image style is the synthetic text paired this image, which is generated via the variable instruction inputted into InternVL~\cite{chen2024internvl}.}
    \scalebox{0.92}{
    \begin{tabular}{l|c|cc}
    \hline\thickhline
    \rowcolor{lightgray}
     &  & \multicolumn{2}{c}{Edited prompt} \\
     \rowcolor{lightgray}
     \multirow{-2}{*}{Editorial object} & \multirow{-2}{*}{Original prompt} & Template & Descriptor list \\
    \midrule
     Background & \emph{a person in the street} & \emph{a person in the \{background\}} & \emph{\{background\} = [desert, beach, airport, farm, lake]} \\
     Weather & \emph{a person on a sunny day} & \emph{a person on a \{weather\} day} & \emph{\{weather\} = [cloudy, rainy, foggy, snowy]} \\
     Style & synthetic text & \emph{\{style\}:} synthetic text & \emph{\{style\} = [oil painting, anime, street photography, van gogh style, pointillism]} \\
     Posture & \emph{a person} & \emph{a person, \{posture\}} & \emph{\{posture\} = [sitting, raising hands, giving a thumbs up, running, walking]} \\
    \hline\thickhline
    \end{tabular}
    }
    \label{SM-tab-1}
\end{table*}

\subsection{Intra-class image augmentation}
To enrich the diversity of synthetic images from the inter-class image generation module, we develop the intra-class image augmentation module. The main goal is to generate multiple person images of the same identities. 

Specifically, we edit the image from the inter-class image generation pipeline by controlling the cross-attention computation in SD model.
This allows us to generate new images while preserving the original identities.
At each denoising step in SD model, cross-attention fuses noisy image embeddings with prompt text to generate spatial attention maps for each textual token.
The attention mechanism is formulated as follows,
\begin{equation}
     \text{Attention}(Q, K, V) = AV = \text{Softmax}(\frac{QK^T}{\sqrt{d}})V,
\end{equation}
where $Q$ is the query embeddings projected from the noisy image, $K$ and $V$ are the key and value embeddings projected from the prompt text, $d$ is the dimension of the query and key embeddings.
$A$ is the cross-attention map where $A_{ij}$ defines the weight of the value of the $j$-th prompt token on the $i$-th image pixel.
We employ inversion technology~\cite{cho2024noise} to generate the initial noise image.

To thoroughly explore synthetic data in TBPR, we examine three key aspects of image editing.

\textbf{Local editing.}
It involves altering a specific element in the image to create an edited image, such as transforming the background in the image from a street to an airport environment.
It is executed by substituting the relevant local textual description in the prompt. For instance, if the original prompt is `\emph{a person in the street}', the edited prompt can be `\emph{a person in the airport}'.
Then the cross-attention map in generating such edited images is computed~\cite{hertz2022prompt} as
\begin{equation}   
    \hat{A} = \begin{cases}
        \hat{A} & t < \tau\\
        A & \text{otherwise},
    \end{cases}
    \label{eq-LE}
\end{equation}
where $\hat{A}$ and $A$ denote the attention maps from generating with the edited and original prompts, respectively. 
It can be seen that we incorporate the attention maps from original image\footnote{For clarity, \emph{the original image} in this section refers to synthetic images from the inter-class generation pipeline, distinguishing them from \emph{the original image} in other sections, which denotes real TBPR benchmark data.} generation into the process of edited image generation after $\tau$-th denoising step. 
It serves to maintain the structure of the original image in generating edited image.

\textbf{Global editing.}
It entails modifying all parts of the image while preserving the original composition, such as transforming the image style from a natural aesthetic to an oil painting.
It is achieved by adding new textual descriptions to the original prompt, such as transforming it into `\emph{oil painting: a person in the street}'.
Then we compute the cross-attention map in generating the edited images~\cite{hertz2022prompt},
\begin{equation}   
    \hat{A}_{ij} = \begin{cases}
        \hat{A}_{ij} & \hat{p}_j \in \text{the added tokens}\\
        A_{ij} & \text{otherwise},
    \end{cases}
\end{equation}
where $\hat{p}_j$ is $j$-th token in the edited prompt.
It can be understood as: the attention values corresponding to the newly added prompt tokens are derived from the new attention map between the edited prompt and the image, while the remaining attention values still retain the original attention map between the original prompt and the image.

\textbf{Non-rigid editing.}
We further consider the non-rigid changes in the person image, such as changes in posture.
Specifically, we augment the prompts by enriching their posture-related textual descriptions, for example, transferring the original prompt from `\emph{a person}' to `\emph{a person, running}'.
Then we perform a mutual cross-attention mechanism~\cite{cao2023masactrl},
\begin{equation}   
    \text{Attention}(\hat{Q}, \hat{K}, \hat{V}) = \begin{cases}
        \text{Attention}(\hat{Q}, K, V) & t < \tau\\
        \text{Attention}(\hat{Q}, \hat{K}, \hat{V}) & \text{otherwise}.
    \end{cases}
    \label{eq-NRE}
\end{equation}
We use the query embeddings $\hat{Q}$ in edited attention mechanism to query the corresponding key and value embeddings $K, V$ in original attention mechanism before $\tau$-th denoising step. 
It essentially utilizes the content from the original image as the foundation for generation, aiding in the preservation of the original structure. Simultaneously, the non-rigid variances are modulated by the edited text prompt.

\begin{figure}[t]
\setlength{\abovecaptionskip}{0.05cm}
\centerline{\includegraphics[width=0.84\linewidth,height=0.35\linewidth]{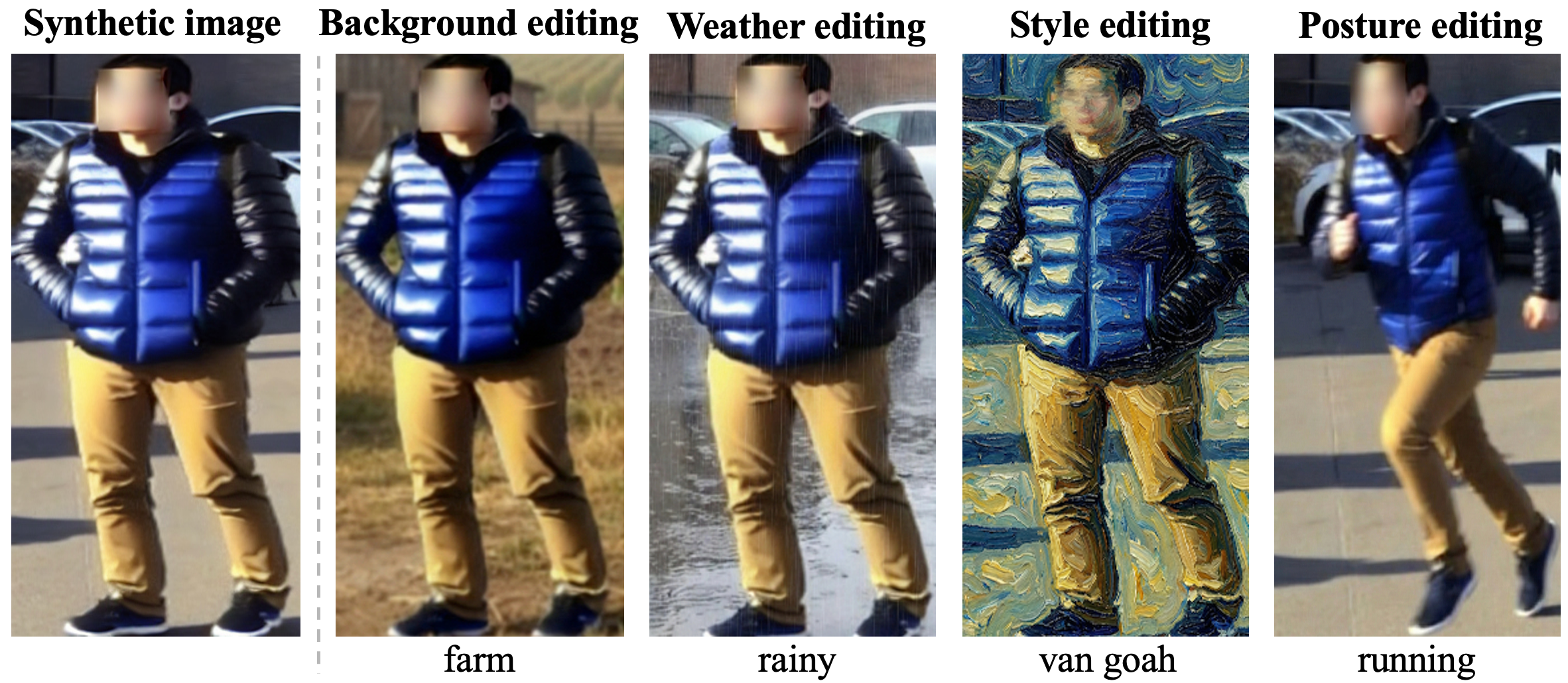}}
\caption{Illustration of edited images under the third scenario. More editing examples are shown in the supplementary material.}
\label{fig-edit}
\end{figure}

Based on the three editing mechanisms, we edit the images along four dimensions: background, weather, style, and posture.
Background editing and weather editing fall under local editing, as they modify specific regions of the image (\emph{i.e.}, background behind the person in the image), style editing is categorized as global editing, as it alters the overall appearance, and posture editing naturally aligns with non-rigid editing.
The prompts used in these image editing processes are specified in Table~\ref{SM-tab-1}.
They are designed empirically, informed by the preferences of text-driven image editing models~\cite{cho2024noise,hertz2022prompt,cao2023masactrl}.
In the image editing process, editing objects and their corresponding descriptor terms are sampled uniformly at random.
An example of edited images is shown in Fig.~\ref{fig-edit} and the distribution across editing conditions is provided in Fig.~\ref{SM-edit-propor}.

\subsection{Text generation}
We leverage the Multimodal Large Language Models (MLLMs)~\cite{chen2024internvl,hu2024minicpm,bai2023qwen} to generate textual descriptions for synthetic person images, eliminating the need for labor-intensive manual annotation.
Specifically, a user-specified instruction is inputted into the MLLMs to guide the generation of descriptions for the given images, with a key technical aspect being the design of effective instructions.
Natively, we provide a constant instruction based on our experience, as follows:

\begin{tcolorbox}[colback=white, colframe=black, opacityframe=0.5]
\small \emph{Don't mention the background environment of the people in the image. Please provide a detailed description of this person's age, gender, top (including color and style), bottom (including color and style), hair (including color and style), shoes (including color and style), and belongings (including color and style). Finally, combine all the details into a single sentence.}
\end{tcolorbox}

Nonetheless, the descriptions generated tend to exhibit a similar structure for each images because of the identical instruction applied to all. The TBPR model trained on such homogeneous descriptions may struggle to generalize effectively to real-world descriptions.
For this, inspired by~\cite{tan2024harnessing}, we design a variable instruction as follows:

\begin{tcolorbox}[colback=white, colframe=black, opacityframe=0.5]
\small \emph{Generate a description about the overall appearance of the person, including the clothing, shoes, hairstyle, gender and belongings,
in a style similar to the template: `\{template\}'. If some requirements in the template are not visible, you can ignore. Do not imagine any contents that are not in the image.}
\end{tcolorbox}

As applied to each image, \emph{\{template\}} in the instruction is randomly replaced from a template pool comprising $121$ templates, obtained by multiple rounds of dialogue and iterative optimization with ChatGPT.
Details about the template pool are provided in the supplementary material.

\begin{table*}
\setlength{\abovecaptionskip}{0.1cm}
    \centering
    \caption{Statistical comparison between our generated datasets and other open-source synthetic datasets for TBPR. These datasets require real person data for synthesis, whereas our SynTBPR series spans the full spectrum of real-data availability, including zero real data.}
    \scalebox{0.92}{
    \begin{tabular}{l|c|cc|c|c|c}
            \toprule
            \rowcolor{lightgray}
            & & \multicolumn{2}{c|}{Synthetic Dataset Overview} &  &  &  \\
            \rowcolor{lightgray}
            \multirow{-2}{*}{Dataset} & \multirow{-2}{*}{Data Sources for Synthesis} & Image & Text & \multirow{-2}{*}{\#Image} & \multirow{-2}{*}{\#Avg Texts/Image} & \multirow{-2}{*}{\#Avg Text Length} \\
            \midrule
            MALS~\cite{yang2023towards} & real texts used ($\sim$120K) & synthetic & synthetic & 1.51M & 1 & 26.96 \\
            LUPerson-T~\cite{shao2023unified} & real images used ($\sim$1.3M) & real & synthetic & 1.30M  & 1 & 27.39 \\
            LUPerson-M~\cite{tan2024harnessing} & real images used ($\sim$1M) & real & synthetic & 1.00M & 4 & 28.61  \\
            HAM-PEDES~\cite{jiang2025modeling} & real images used ($\sim$1M) & real & synthetic & 1.00M  & 2 & 26.30 \\
            \hdashline
            SynTBPR-S1 & no real data used & synthetic & synthetic & 1.60M  & 4  &37.36  \\
            SynTBPR-S2 & few real images used ($\sim$8) & synthetic & synthetic & 1.63M  & 4  &34.74  \\
            SynTBPR-S3 & real images used ($\sim$15K-34K) & synthetic & synthetic & 1.63M  & 4 & 34.10  \\
            \bottomrule
        \end{tabular}
    }
    \label{SM-tab-0}
\end{table*}

\begin{figure}
\setlength{\abovecaptionskip}{0.05cm}
\centerline{\includegraphics[width=0.91\linewidth,height=0.3\linewidth]{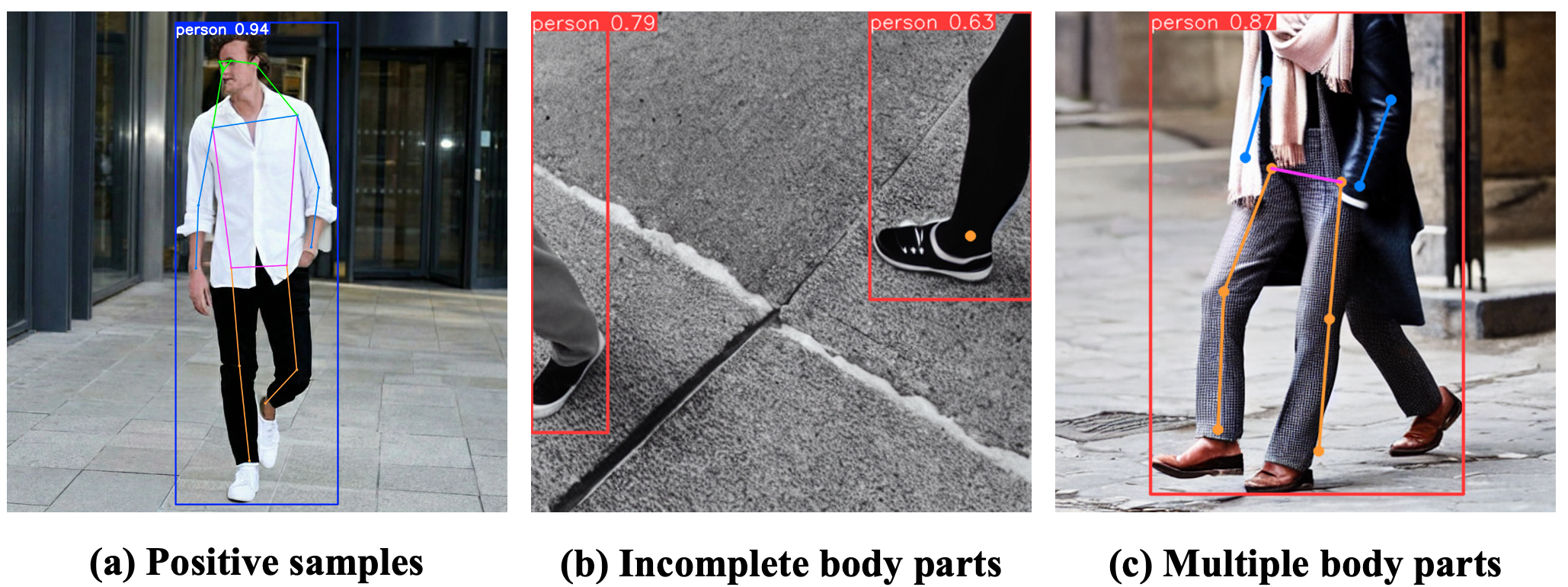}}
\caption{Illustration of a positive image (a) and negative noisy images (b)$\sim$(c) (uncropped). Additional examples are provided in the supplementary material.}
\label{fig-yolo}
\end{figure}

\begin{figure}[t]
\setlength{\abovecaptionskip}{0.05cm}
\centerline{\includegraphics[width=1\linewidth,height=0.9\linewidth]{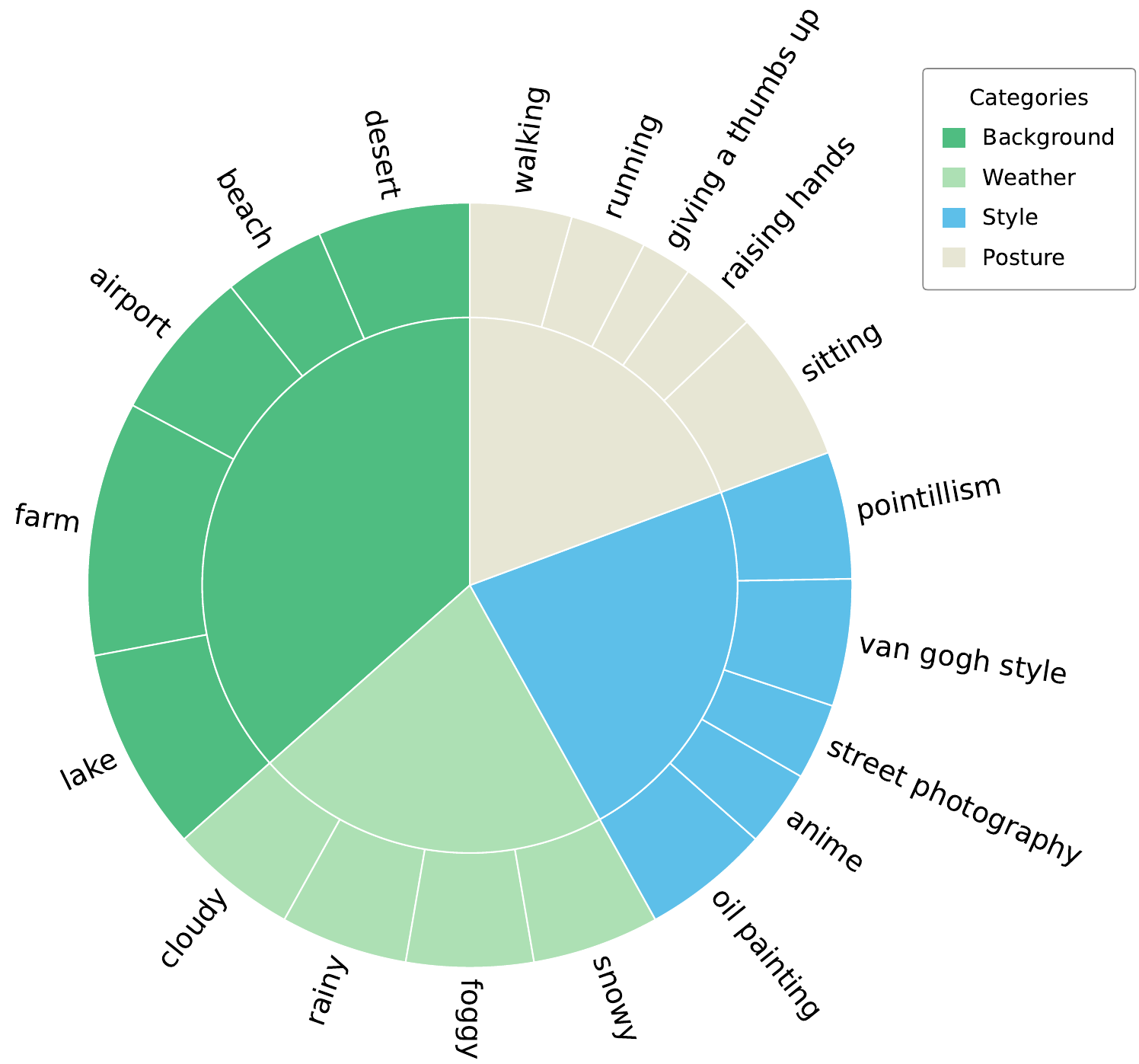}}
\caption{Illustration of the proportion of edited images under each editing condition. The distribution appears approximately uniform, as the editing conditions are sampled uniformly at random during image editing.}
\label{SM-edit-propor}
\end{figure}

\textbf{Data denoising.}
Given that synthetic image and text data inherently contain noise, we consider data denoising before feeding them into the baseline model.
For image data, noise typically arises from incomplete (like missing legs) or multiple body parts (like two heads), so we evaluate the completeness and rationality of human keypoints detected by YOLOv8~\cite{reis2023real} in synthetic person images to filter out noisy images (Fig.~\ref{fig-yolo}).
For text data, noise includes inappropriate symbols (\emph{e.g.,} \emph{`[]'}) or irrelevant information (\emph{e.g.,} \emph{`the image is blurry, ...'}), which we detect and remove.
Detailed denoising results are visualized in the supplementary material.

%



\subsection{Baseline}

To efficiently assess the influence of synthetic data on TBPR performance, we construct a lightweight baseline model.
That is, we adopt the simplified TBPS-CLIP~\cite{cao2024empirical} as our baseline model, with two modifications: one to enhance performance and another to adapt it to synthetic data training. 
Compared with other TBPR models~\cite{bai2023rasa,jiang2023cross}, TBPS-CLIP inherits CLIP's features with a lightweight design, enabling efficient training and inference while keeping competitive performance, making it well-suited for comprehensive data generation studies.

We introduce two modifications:
(1) replacing CLIP backbone with EVA-CLIP~\cite{sun2023eva} for improved efficiency and effectiveness;
(2) exploring three noise-robust learning strategies to enhance synthetic data training: 
label smoothing~\cite{cao2024empirical} to reduce noise impact during loss computation, a two-component Gaussian Mixture Model (GMM)~\cite{qin2024noisy} to identify and exclude noisy data, channel masking~\cite{gao2024semi} to mitigate noise effects. 
The effectiveness of these strategies is evaluated in our experiments.

%% file: sec/4_experiment.tex
\section{Experiments}


\textbf{Datasets}.
We conduct experimental studies to evaluate the effectiveness of synthetic data in TBPR on three datasets.
\textbf{CUHK-PEDES}~\cite{li2017person} comprises $40,206$ images paired with $80,440$ texts from a total of $13,003$ identities. This dataset is divided into $34,054$ images with $68,126$ texts from $11,003$ identities in the training set, $3,078$ images with $6,158$ texts from $1,000$ identities in the validation set, and $3,074$ images with $6,156$ texts from $1,000$ identities in the test set. 
The average length of all texts is $23$.
\textbf{ICFG-PEDES}~\cite{ding2021semantically} contains $54,522$ images from $4,102$ identities, all captured in a single environment. Each image is paired with a single, highly descriptive text caption—averaging $37$ words—making it more detailed than CUHK-PEDES. The dataset is split into $34,674$ images from $3,102$ identities in the training set and $19,848$ images from $1,000$ identities in the test set.
\textbf{RSTPReid}~\cite{zhu2021dssl} comprises $20,505 $ images of $4,101$ identities with $5$ images per identity captured from different camera views. Each image is annotated with $2$ textual descriptions, each containing at least $23$ words. The dataset splits $3,701$ identities for training, $200$ for validation, and $200$ for testing.

Also, we conduct experiments on our synthetic datasets. Table~\ref{SM-tab-0} presents a statistical comparison between our datasets (SynTBPR-S1, SynTBPR-S2 and SynTBPR-S3) generated under three scenarios, with other open-source synthetic datasets.

\textbf{Evaluation metrics}.
We employ the widely used Rank-k metric ($k=1, 10$) and the mean average precision (mAP) to assess performance. 
Given a query text, all test images are ranked by their similarity to the query. A search is deemed successful if the top-$k$ images include any matching identity. 
Higher Rank-k and mAP scores indicate better performance.

\textbf{Implementation details}.
All experiments are conducted on NVIDIA H800 GPUs. 
\ding{182} For image generation, we leverage Stable Diffusion-v1.5 (SD-v1.5) guided by $50,000$ prompts.
These prompts are derived from our proposed five types of description templates, with each type generating $10,000$ prompts. The guidance scale in SD-v1.5 is set to $8.5$.
In the first scenario, we adopt $50,000$ prompts with different random seeds to guide SD-v1.5 for generating images.
In the second scenario, we train SD-v1.5 with input image resizing to $512 \times 512$, $\lambda=1$ in Eq.~\ref{twoS-loss}, a learning rate of $2e-6$, and 800 iterations.
In the third scenario, we train SD-v1.5 with LoRA (rank $=6$), where all parameters of SD-v1.5 are fixed, and new parameters from LoRA and text encoder are fine-tuned.
Training is performed with a batch size of $64$ for $4,000$ iterations.
The initial learning rate is $1.0$ and subsequently is automatically adjusted via the prodigy optimizer~\cite{mishchenko2023prodigy}.
\ding{183} For image augmentation, 
$\tau$ in Eq.~\ref{eq-LE} for local editing is configured to be set at $80\%$ of all denoising steps, while $\tau$ in Eq.~\ref{eq-NRE} for non-rigid editing is set to $4$.
\ding{184} For text generation, the constant instruction is inputted into InternVL~\cite{chen2024internvl}, while the variable instruction is fed to InternVL, MiniCPM~\cite{hu2024minicpm}, and QwenVL~\cite{bai2023qwen}, respectively.
This results in four structurally diverse textual descriptions for each image.
\ding{185} For TBPR baseline model, the generation model outputs full-view images at a resolution of $512\times512$, based on which we perform object detection using YOLOv8~\cite{reis2023real} to crop person images and are then used as input for the TBPR baseline model. Input images are resized to $224 \times $224, while texts with a fixed length of $77$ are input into the model. 
For the TBPR baseline model, the generation model produces full-view images at a resolution of $512\times512$, based on which we perform object detection using YOLOv8~\cite{reis2023real} to crop person regions. The cropped person images are resized to $224 \times $224, and the corresponding texts, padded or truncated to a fixed length of $77$ tokens, are provided as input to the retrieval model.
EVA-CLIP~\cite{sun2023eva}, as the retrieval model, is initialized with its public pretrained weights.
We use the AdamW optimizer with a linear warm-up followed by cosine decay.
The linear warm-up starts from $1e-6$ and gradually increases to $5e-6$. 
During training with synthetic data, we set the initial learning rate (after warm-up) to $3e-5$ and train for $8$ epochs;
for real data, the initial learning rate is set to $1e-4$ with $5$ epochs of training.

\begin{table*}
\setlength{\abovecaptionskip}{0.05cm}
    \centering
    \caption{Study on the effectiveness of synthetic data from the inter-class image generation (gen.) and intra-class image augmentation (aug.) across three scenarios (S1: No data, S2: Limited data, S3: Abundant data). Results are obtained by jointly training on both real and synthetic data.
    Synthetic data (million, M) is derived based on real data marked with same color. The `all' refers to all training data in the respective dataset. In S2, the baseline with $8$ real pairs fails to converge, so we present the same results as the baseline in S1.} 
    \scalebox{0.86}{
    \begin{tabular}{l|l c|cc|ccc|ccc|ccc|cc}
    \toprule
    \rowcolor{lightgray}
     &  &  & \multicolumn{2}{c|}{\# Synthetic data} & \multicolumn{3}{c|}{CUHK-PEDES} & \multicolumn{3}{c|}{ICFG-PEDES} & \multicolumn{3}{c|}{RSTPReid} & \multicolumn{2}{c}{Average Gain}\\
    \cline{6-16}
    \rowcolor{lightgray}
    \multirow{-2}{*}{Method} & \multicolumn{2}{c|}{\multirow{-2}{*}{\# Real data}} & Gen. & Aug. & Rank-1 & Rank-10 & mAP & Rank-1 & Rank-10 & mAP & Rank-1 & Rank-10 & mAP & Rank-1 & mAP\\
    \midrule
    Baseline &  &  & 0 & 0 & 22.27 & 49.37  & 20.20  & 12.34  & 32.52  & 5.00  & 20.00 & 57.15   & 16.19 & -& - \\
    + Ours (gen.) &  &  & \textcolor[rgb]{0.6,0.5,0.7}{\textbf{1.5M}} & 0 & 54.30 & 79.53  & 48.92  &38.92 &64.05   & 21.22   &45.10   &76.80    & 36.06 & 31.24 & 21.60 \\
    + Ours (gen.\&aug.) &  &  & \textcolor[rgb]{0.6,0.5,0.7}{\textbf{1.5M}}  & \textcolor[rgb]{0.6,0.5,0.7}{\textbf{0.1M}} &54.87   &80.30    &49.58    &39.40   &64.34   &21.29   &45.20   &77.20  &36.42 & 31.61 & 21.97 \\
    \cdashline{1-1}\cdashline{4-16}
    IRRA~\cite{jiang2023cross} &  &  & 0 & 0 & 12.61 & 35.52 & 11.14 & 6.67 & 24.89 & 2.52 & 13.45 & 42.20 & 10.30 & - & - \\
    + Ours (gen.) &  &  & \textcolor[rgb]{0.6,0.5,0.7}{\textbf{1.5M}} & 0 & 50.03 & 77.87 & 45.23 & 34.88 & 61.92 & 18.23 & 42.55 & 74.80 & 32.24 & 31.58 & 23.91 \\
    + Ours (gen.\&aug.) &  &  & \textcolor[rgb]{0.6,0.5,0.7}{\textbf{1.5M}}  & \textcolor[rgb]{0.6,0.5,0.7}{\textbf{0.1M}} & 51.88 & 79.01 & 47.22 & 35.34 & 63.02 & 19.05 & 43.17 & 76.00 & 34.41 & 32.55 & 25.57 \\
    \cdashline{1-1}\cdashline{4-16}
    RaSa~\cite{bai2023rasa} &  &  & 0 & 0 & 27.27 & 49.56 & 20.19 & 14.06 & 33.06 & 4.83 & 26.75 & 62.10 & 18.12 & - & - \\
    + Ours (gen.) &  &  & \textcolor[rgb]{0.6,0.5,0.7}{\textbf{1.5M}} & 0 & 57.23 & 80.97 & 50.11 & 41.02 & 66.89 & 21.28 & 46.32 & 76.88 & 36.75 & 25.48 & 21.63 \\
    + Ours (gen.\&aug.) & \multirow{-9}{*}{\textbf{\textcolor[rgb]{0.6,0.5,0.7}{S1}}:} & \multirow{-9}{*}{0} & \textcolor[rgb]{0.6,0.5,0.7}{\textbf{1.5M}}  & \textcolor[rgb]{0.6,0.5,0.7}{\textbf{0.1M}} & 57.82 & 81.03 & 50.23 & 41.67 & 66.97 & 21.49 & 46.41 & 76.76 & 37.03 & 25.97 & 21.87 \\
    \hline
    Baseline &  &  & 0 & 0  & 22.27  & 49.37  & 20.20  & 12.34  & 32.52  & 5.00  & 20.00  & 57.15   & 16.19 & - & - \\
    + Ours (gen.) &  &  & \textcolor[rgb]{0.6,0.6,0.3}{\textbf{0.1M}} & 0 & 54.11   &79.24   & 48.23  & 39.89 &   64.41   & 20.45 & 42.70  &73.35    & 32.51 & 30.70 &  19.93 \\
    + Ours (gen.) &  &  & \textcolor[rgb]{0.6,0.5,0.7}{\textbf{1.5M +}} \textcolor[rgb]{0.6,0.6,0.3}{\textbf{0.1M}} & 0 & 55.05   &80.54   &49.61   & 40.47   & 65.28  & 21.77 & 44.90    &75.50    & 34.69   & 31.94 & 21.56 \\
    + Ours (gen.\&aug.) &  &  & \textcolor[rgb]{0.6,0.5,0.7}{\textbf{1.5M +}} \textcolor[rgb]{0.6,0.6,0.3}{\textbf{0.1M}} & \textcolor[rgb]{0.6,0.6,0.3}{\textbf{0.03M}}  &55.28   &81.09   & 49.72  &40.76   & 65.38   & 21.68  &  46.65 & 77.40   & 35.34 & 32.69 & 21.78 \\
    \cdashline{1-1}\cdashline{4-16}
    IRRA~\cite{jiang2023cross} &  &  & 0 & 0 & 12.61 & 35.52 & 11.14 & 6.67 & 24.89 & 2.52 & 13.45 & 42.20 & 10.30 & - & - \\
    + Ours (gen.) &  &  & \textcolor[rgb]{0.6,0.6,0.3}{\textbf{0.1M}} & 0 & 50.07 & 78.43 & 45.55 & 35.88 & 62.78 & 18.66 & 43.23 & 75.73 & 33.79 & 32.15 & 24.68 \\
    + Ours (gen.) &  &  & \textcolor[rgb]{0.6,0.5,0.7}{\textbf{1.5M +}} \textcolor[rgb]{0.6,0.6,0.3}{\textbf{0.1M}} & 0 & 52.07 & 80.22 & 48.21 & 36.17 & 62.89 & 18.97 & 43.79 & 76.53 & 34.42 & 33.10 & 25.88 \\
    + Ours (gen.\&aug.) &  &  & \textcolor[rgb]{0.6,0.5,0.7}{\textbf{1.5M +}} \textcolor[rgb]{0.6,0.6,0.3}{\textbf{0.1M}} & \textcolor[rgb]{0.6,0.6,0.3}{\textbf{0.03M}} & 52.89 & 80.43 & 48.33 & 37.02 & 63.28 & 19.23 & 44.52 & 76.98 & 34.76 & 33.57 & 26.12 \\
    \cdashline{1-1}\cdashline{4-16}
    RaSa~\cite{bai2023rasa} &  &  & 0 & 0 & 27.27 & 49.56 & 20.19 & 14.06 & 33.06 & 4.83 & 26.75 & 62.10 & 18.12 & - & - \\
    + Ours (gen.) &  &  & \textcolor[rgb]{0.6,0.6,0.3}{\textbf{0.1M}} & 0 & 57.01 & 81.00 & 50.13 & 41.55 & 66.74 & 21.17 & 46.89 & 76.78 & 37.01 & 25.79 & 21.72 \\
    + Ours (gen.) &  &  & \textcolor[rgb]{0.6,0.5,0.7}{\textbf{1.5M +}} \textcolor[rgb]{0.6,0.6,0.3}{\textbf{0.1M}} & 0 & 58.13 & 82.04 & 51.22 & 42.23 & 67.12 & 22.05 & 47.35 & 77.12 & 37.19 & 26.54 & 22.41 \\
    + Ours (gen.\&aug.) & \multirow{-12}{*}{\textcolor[rgb]{0.6,0.6,0.3}{\textbf{S2}}:} &  \multirow{-12}{*}{8} & \textcolor[rgb]{0.6,0.5,0.7}{\textbf{1.5M +}} \textcolor[rgb]{0.6,0.6,0.3}{\textbf{0.1M}} & \textcolor[rgb]{0.6,0.6,0.3}{\textbf{0.03M}} & 58.23 & 82.69 & 51.98 & 42.45 & 67.88 & 22.23 & 47.65 & 77.89 & 37.63 & 26.74 & 22.89 \\
     \hline
    Baseline &  &  & 0 & 0 & 73.51  & 92.95  & 65.74  & 65.85 &  86.04  & 40.73 & 61.35  & 87.80  & 48.28 & - & - \\
    + Ours (gen.) &  &  & \textcolor[rgb]{0.4,0.5,0.6}{\textbf{0.1M}} & 0 &73.80   & 93.05    &66.58    & 66.02 & 86.00   &42.19  & 63.75  & 89.25    &50.87   & 0.95 & 1.63 \\
    + Ours (gen.) &  &  & \textcolor[rgb]{0.6,0.5,0.7}{\textbf{1.5M}} + \textcolor[rgb]{0.4,0.5,0.6}{\textbf{0.1M}} & 0 & 75.47 & 93.68   & 67.89  &67.28  &  86.84    & 42.98  &68.00   & 91.25     & 53.46    & 3.35 & 3.20 \\
    + Ours (gen.\&aug.) &  &  & \textcolor[rgb]{0.6,0.5,0.7}{\textbf{1.5M +}} \textcolor[rgb]{0.4,0.5,0.6}{\textbf{0.1M}} & \textcolor[rgb]{0.4,0.5,0.6}{\textbf{0.03M}} &76.01  &93.89    &67.98    &67.62  &86.76    & 43.10    &68.65  &90.65   &53.42  & 3.86 & 3.25 \\
    \cdashline{1-1}\cdashline{4-16}
    IRRA~\cite{jiang2023cross} &  &  & 0 & 0 & 73.38 & 93.71 & 66.13 & 63.46 & 85.82 & 38.06 & 60.20 & 88.20 & 48.51 & - & - \\
    + Ours (gen.) &  &  & \textcolor[rgb]{0.4,0.5,0.6}{\textbf{0.1M}} & 0 & 73.87 & 93.86 & 66.72 & 65.23 & 86.11 & 39.77 & 65.22 & 88.21 & 48.47 & 2.43 & 0.75 \\
    + Ours (gen.) &  &  & \textcolor[rgb]{0.6,0.5,0.7}{\textbf{1.5M}} + \textcolor[rgb]{0.4,0.5,0.6}{\textbf{0.1M}} & 0 & 75.99 & 94.21 & 68.02 & 67.01 & 86.76 & 40.23 & 67.97 & 90.35 & 52.23 & 5.64 & 2.59 \\
    + Ours (gen.\&aug.) &  &  & \textcolor[rgb]{0.6,0.5,0.7}{\textbf{1.5M}} + \textcolor[rgb]{0.4,0.5,0.6}{\textbf{0.1M}} & \textcolor[rgb]{0.4,0.5,0.6}{\textbf{0.03M}} & 76.32 & 94.51 & 68.78 & 67.30 & 87.12 & 41.61 & 68.40 & 90.80 & 53.05 & 5.00 & 3.58 \\
    \cdashline{1-1}\cdashline{4-16}
    RaSa~\cite{bai2023rasa} &  &  & 0 & 0 & 76.51 & 94.25 & 69.38 & 65.28 & 85.12 & 41.29 & 66.90 & 91.35 & 52.31 & - & - \\
    + Ours (gen.) &  &  & \textcolor[rgb]{0.4,0.5,0.6}{\textbf{0.1M}} & 0 & 76.76 & 94.21 & 70.02 & 66.17 & 85.88 & 42.01 & 67.23 & 91.41 & 53.09 & 0.49 & 0.71 \\
    + Ours (gen.) &  &  & \textcolor[rgb]{0.6,0.5,0.7}{\textbf{1.5M}} + \textcolor[rgb]{0.4,0.5,0.6}{\textbf{0.1M}} & 0 & 78.02 & 94.55 & 71.23 & 68.99 & 87.07 & 44.87 & 69.08 & 91.90 & 55.25 & 3.13 & 2.82 \\
    + Ours (gen.\&aug.) & \multirow{-12}{*}{\textcolor[rgb]{0.4,0.5,0.6}{\textbf{S3}}:} & \multirow{-12}{*}{all} & \textcolor[rgb]{0.6,0.5,0.7}{\textbf{1.5M}} + \textcolor[rgb]{0.4,0.5,0.6}{\textbf{0.1M}} & \textcolor[rgb]{0.4,0.5,0.6}{\textbf{0.03M}} & 78.65 & 94.69 & 71.54 & 69.63 & 87.41 & 45.48 & 70.75 & 92.20 & 55.74 & 3.45 & 3.26 \\
    \bottomrule
    \end{tabular}
    }
    \label{tab-1.1}
\end{table*}

\subsection{Evaluating the effectiveness of synthetic data}
Based on our constructed baseline, and complemented by two widely used TBPR baselines (IRRA~\cite{jiang2023cross} and RaSa~\cite{bai2023rasa}), we conduct the empirical studies on synthetic data from the proposed inter-class image generation (\emph{Ours (gen.)}) and intra-class image augmentation (\emph{Ours (aug.)}) across three scenarios: no data (S1), limited data (S2), and abundant data (S3).

\textbf{Study on inter-class image generation.} 
The results are shown in Table~\ref{tab-1.1}.
(1) In S1, integrating \emph{Ours (gen.)} into \emph{Baseline} significantly boosts performance across all datasets. 
Our pipeline generates large-scale synthetic data ($1.5$ million here) to enrich the baseline with domain knowledge, enhancing performance.
(2) In S2, the limited real data fails to train baseline, degrading same results with in S1. Fortunately, our pipeline can generate abundant synthetic data based on the $8$ real pairs\footnote{The 8 pairs are randomly sampled from the training set and serve to provide domain-specific contextual cues about pedestrian style—sufficient to ensure robust performance.}, which are then employed to train baseline, thereby improving performance. 
Further, pre-training with $1.5$ million synthetic samples from S1 further enhances results.
(3) In S3, a startup promising performance has been achieved due to the abundance of real data for training baseline. However, incorporating \emph{Ours (gen.)} further enhances results.
(4) Compared to state-of-the-art methods (in Table~\ref{tab-1}), which rely on real data for training, our data generation pipeline integrated into the constructed baseline achieves promising results.

\begin{figure*}
\setlength{\abovecaptionskip}{0.15cm}
\centerline{\includegraphics[width=0.95\linewidth,height=0.19\linewidth]{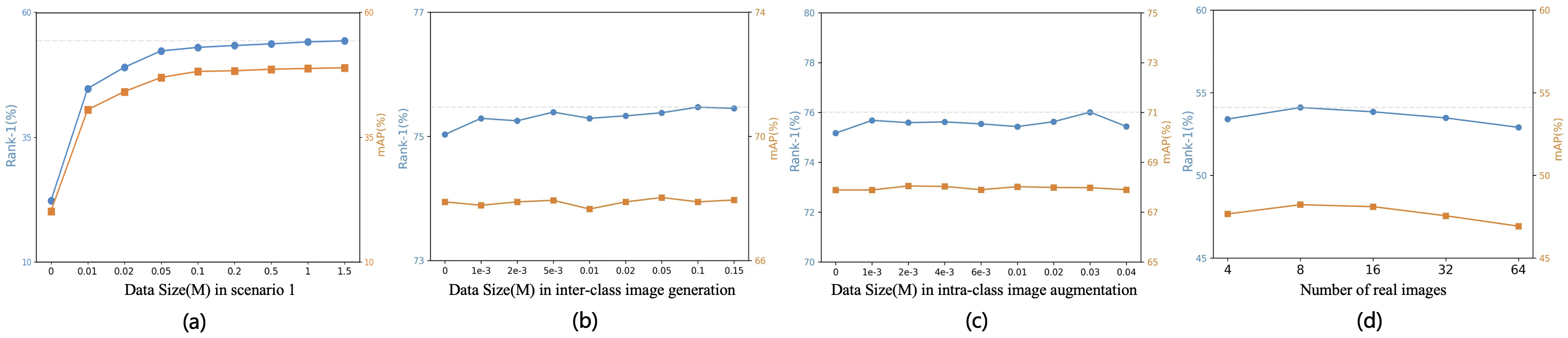}}
\caption{Performance trend under different number of synthetic data and different number of real data on CUHK-PEDES.}
\label{SM-number}
\end{figure*}





\textbf{Study on intra-class image augmentation.}
As shown in Table~\ref{tab-1.1}, adding synthetic data ($0.03$ million) from our intra-class image augmentation (\emph{Ours (gen.\&aug.)}) consistently enhances performance across baselines on all datasets. 

\textbf{Analysis of variable parameter in image generation and augmentation.}
In our proposed pipeline, the number of synthetic data can vary. 
Specifically, we generate $1.5$ million synthetic samples in S1, $0.1$ million in S2 and S3 using the inter-class image generation module, and an additional $0.03$ million via the intra-class image augmentation module in our experiments.
Furthermore, we set the real data to $8$ pairs to simulate S2, though this parameter can also vary.
To assess the impact of these parameters, we evaluate performance on CUHK-PEDES with varying values using our constructed baseline.
(1) Effect of synthetic data in S1. Results for varying synthetic data amounts in S1 are in Fig.~\ref{SM-number} (a). Performance improves continuously with more synthetic data, then stabilizes.
(2) Effect of synthetic data in S3.
Taking Scenario 3 as an example, Fig.~\ref{SM-number} (b) and (c) show performance initially improves slightly but then declines with more synthetic data. 
This suggests synthetic data adds only marginal gains to large real datasets due to the quality gap—and too much can even hurt performance.
(3) Effect of the real data in S2. As shown in Fig.~\ref{SM-number} (d), the performance first improves then declines as real data increases. Our training loss functions in Eq.~\ref{twoS-loss} are designed to learn representations for specific pedestrian objects based on a robust SD model capable of generating new images. In this context, a small amount of data suffices to optimize Eq.~\ref{twoS-loss} for representation modeling, while slightly more can make it harder.

\begin{table*}
    \centering
    \caption{Comparisons with state-of-the-art methods. The method marked `$*$' indicates reproduced results using the published code$^3$. For ease of comparison, we report the average results across the three datasets.}
    \scalebox{0.86}{
    \begin{tabular}{l|c|ccc|ccc|ccc|ccc}
    \toprule
    \rowcolor{gray!25}
     & & \multicolumn{3}{c|}{CUHK-PEDES} & \multicolumn{3}{c|}{ICFG-PEDES} & \multicolumn{3}{c|}{RSTPReid} & \multicolumn{3}{c}{Average} \\
    \rowcolor{gray!25}
    \multirow{-2}{*}{Method}  & \multirow{-2}{*}{Reference}  & Rank-1 & Rank-10 & mAP & Rank-1 & Rank-10 & mAP & Rank-1 & Rank-10 & mAP & Rank-1 & Rank-10 & mAP \\
    \midrule
    \multicolumn{14}{l}{\textit{Comparison with methods without synthesizing data:}}\\
    \hline
    ViTAA~\cite{wang2020vitaa} & ECCV20  & 55.97 & 83.52 & - & - & - & - & - & - & -  & 55.97 & 83.52 & - \\
    LapsCore~\cite{wu2021lapscore} & ICCV21 & 63.40 & 87.80 & - & - & - & - & - & - & - & 63.40 & 87.80 & - \\
    IVT~\cite{shu2022see} & ECCV22 & 65.59 & 89.21 & - & 56.04 & 80.22 & - & 46.70 & 78.80 & - & 56.11 & 82.74 & - \\
    SAF~\cite{li2022learning} & ICASSP22 & 64.13 & 88.40 & 58.61 & 54.86 & 79.13 & 32.76 & 44.05 & 76.25 & 36.81 & 54.35 & 81.26 & 42.73 \\
    IRRA~\cite{jiang2023cross} & CVPR23  & 73.38 & 93.71 & 66.13 & 63.46 & 85.82 & 38.06 & 60.20 & 88.20 & 47.17  & 65.68 & 89.24 & 50.45  \\
    BiLMa~\cite{fujii2023bilma} & ICCV23  & 74.03 & 93.62 & 66.57 & 63.83 & 85.74 & 38.26 & 61.20 & 88.80 & 48.51 & 66.35 & 89.39 & 51.11 \\
    RaSa~\cite{bai2023rasa} & IJCAI23 & 76.51 & 94.25 & 69.38 & 65.28 & 85.12 & 41.29 & 66.90 & 91.35 & 52.31 & 69.56 & 90.24 & 54.33 \\
    LAIP~\cite{wu2024laip} & ICME24  & 76.72 & 93.60 & 66.05 & 63.52 & 84.57 & 37.02 & 62.00 & 88.50 & 45.27 & 67.41 & 88.89 & 49.45 \\
    FSRL+Pre~\cite{wang2024fine} & ICMR24  & 74.86 & 94.14 & 67.57 & 64.93 & 86.19 & 40.67 & 60.65 & 89.6 & 48.18 & 66.81 & 89.98 & 52.14 \\
    IRLT~\cite{liu2024causality} & AAAI24  & 74.46 & 94.01 & - & 64.72 & 86.31 & - & 61.49 & 89.23 & - & 66.89 & 89.85 & - \\
    AUL$^*$~\cite{li2024adaptive} & AAAI24 & 75.47  & 93.89  & 66.33  & 67.01  & 87.19  & 38.69  & 67.60  & 90.40  & 51.84  & 70.03 & 90.49 & 52.29 \\
    SAMC~\cite{lu2024mind}  & TIFS24 & 74.03 & 93.31 & 68.42 & 63.68 & 85.21 & 42.41 & 60.80 & 89.00 & 49.67 &	66.17 & 89.17 & 53.50 \\
    LSPM~\cite{li2024learning} & TMM24 & 74.38 & 93.42 & 67.74 & 64.4 & 85.41 & 42.6 & -	& -	& -	& 69.39	& 89.42	& 55.17 \\
    CADA-L~\cite{lin2024cross} & TMM24  & \textbf{78.37} & 94.58 & 68.87 & 67.81 & 87.14 & 39.85 & 69.60 & 92.40 & 52.74 & 71.93 & 91.37 & 53.82 \\
    GAHR~\cite{qi2025granularity} & TIFS25 & 76.64 & 94.10 & 66.81 & 68.69 & 87.40 & 42.10 & 68.85 & 91.10 & 53.60 & 71.39 & 90.87 & 54.17 \\
    CAMeL~\cite{yu2025camel} & TIFS25 & 77.24 & \textbf{95.16} & 68.32 & 68.70 & \textbf{88.32} & 41.58 & 68.50 & \textbf{92.70} & 53.61 & 71.48 & \textbf{92.06} & 54.50 \\
    ICL~\cite{qin2025human} & CVPR25 & 78.18 & 94.83 & \textbf{69.58} & \textbf{69.22} & 88.06 & \textbf{42.34} & \textbf{70.00} & 91.70 & \textbf{54.16} & \textbf{72.47} & 91.53 & \textbf{55.36} \\
    \hline
    \multicolumn{14}{l}{\textit{Comparison with methods with synthesizing data:}}\\
    \hline
    IRRA~\cite{jiang2023cross} & CVPR23 & 73.38 & 93.71 & 66.13 & 63.46 & 85.82 & 38.06 & 60.20 & 88.20 & 47.17 & 65.68 & 89.24 & 50.45 \\
    + MALS~\cite{yang2023towards} & MM23 & 74.05  & 93.64  &66.57  & 64.37 &  86.12 &38.85  & 61.90  &89.30  & 48.08  & 66.77 & 89.69 & 51.17 \\
    + LUPerson-T~\cite{shao2023unified} & ICCV23  &74.37  &93.97  &66.60  &64.50  & 85.74 & 38.22 & 62.20 & 89.75   &48.33  & 67.02 & 89.82 & 51.05 \\
    + LUPerson-M~\cite{tan2024harnessing} & CVPR24  &76.82  & 94.46 &69.55 &67.05 &87.33  &41.51 & 68.50 & 92.10 &53.02  & 70.79 & 91.30 & 54.69 \\
    + HAM-PEDES~\cite{jiang2025modeling} & CVPR25 & 77.71  & 94.57 & 69.68  & 68.25 & 88.15 & 42.30  & 71.69  & \textbf{93.30}  & 55.19  &  72.55 & 92.01 & 55.72 \\
    + SynTBPR-S3 & Ours  &76.32   &94.51 & 68.78 &67.30 &87.12   &41.61  & 68.40 &90.80   &53.05   & 70.67 & 90.81 & 54.48 \\
    \hdashline
    RaSa~\cite{bai2023rasa} & IJCAI23 & 76.51 & 94.25 & 69.38 & 65.28 & 85.12 & 41.29 & 66.90 & 91.35 & 52.31 &  69.56 & 90.24 & 54.33 \\
    + MALS~\cite{yang2023towards} & MM23  &76.38  & 94.28  &69.38 &66.21  & 85.34  &41.93  &69.50   & 92.85  &54.09  & 70.70 & 90.82 & 55.13 \\
    + LUPerson-T~\cite{shao2023unified} & ICCV23  &75.78    & 93.83  &69.07   &65.64  &85.39 &41.58  & 68.00  &90.50   &53.28  & 69.81 & 89.91 & 54.64 \\
    + LUPerson-M~\cite{tan2024harnessing} & CVPR24  &78.22 & 94.96   &71.12  &68.91 &87.57  &44.74 & 70.60  & 92.90   &\textbf{56.52}  & 72.58 & 91.81 & 57.46 \\
    + HAM-PEDES~\cite{jiang2025modeling} & CVPR25 & \textbf{79.01}   & 94.70 & \textbf{72.39}  & 70.30  & 87.81  & \textbf{47.58}  & 69.75 & 90.20  & 56.35 & 73.02 & 90.90 & \textbf{58.77} \\
    + SynTBPR-S3 & Ours  &78.65   &94.69 & 71.54  &69.63  &87.41   & 45.48  & 70.75   & 92.20  &55.74   & 73.01 & 91.43 & 57.59 \\
    \hdashline
    AUL$^*$~\cite{li2024adaptive} & AAAI24 & 75.47  & 93.89  & 66.33  & 67.01  & 87.19  & 38.69  & 67.60  & 90.40  & 51.84  &  70.03 & 90.50 & 52.29\\
    + MALS~\cite{yang2023towards} & MM23  &74.87   &94.02 &65.94  &66.51  & 86.68  & 38.20  &66.70 &90.50  & 51.28 & 69.36 & 90.40 & 51.81 \\
    + LUPerson-T~\cite{shao2023unified} & ICCV23  &75.63  &94.36  &66.22  &66.62   &87.19 &38.46  & 67.55  &90.70   & 51.93 & 69.93 & 90.75 & 52.20  \\
    + LUPerson-M~\cite{tan2024harnessing} & CVPR24 & 78.80  & 94.93  &69.26   & 69.60  & 88.33  &41.20   &70.35   & 91.45  & 54.03 &  72.92 & 91.57 & 54.83 \\
    + HAM-PEDES~\cite{jiang2025modeling} & CVPR25 & 77.86  & \textbf{95.22}  & 69.01  & \textbf{70.77}  & \textbf{89.10}  & 40.73  & \textbf{71.90}  & 91.75  & 55.51 & \textbf{73.51} & \textbf{92.02} & 55.08 \\
    + SynTBPR-S3 & Ours & 78.72  & 94.83  & 68.96  & 70.38  & 88.74  & 41.20  & 70.55  & 91.95 & 52.58  & 73.22 & 91.84 & 54.25 \\
    \bottomrule
    \end{tabular}
    }
    \label{tab-1}
\end{table*}

\subsection{Comparisons with state-of-the-art methods}

\begin{table}
\setlength{\abovecaptionskip}{0.05cm}
    \centering
    \caption{Ablation results of adopting different image generation method on CUHK-PEDES. FID and Diversity (Div.) are two metrics for evaluating the quality of synthetic images.}
    \scalebox{0.95}{
    \begin{tabular}{l|cccc}
    \toprule
    \rowcolor{gray!25}
    Method &  Rank-1 & mAP & FID($\downarrow$) & Div.($\uparrow$) \\
    \midrule
    Baseline  &22.27   & 20.20  & - & - \\
    \hline
    + Ours (gen.)~\cite{rombach2022high}  & 50.81   & 45.50  & 137.00 & 0.751 \\
    + Ours (gen.)~\cite{chen2023pixart}  & 45.31     & 40.65  & 189.76 & 0.733 \\
    \hline
    + Ours (gen.\&aug.)~\cite{cao2023masactrl,hertz2022prompt} & 54.68    &49.30   & 121.74 & 0.760 \\
    + Ours (gen.\&aug.)~\cite{cao2023masactrl,brooks2023instructpix2pix} &52.68   & 47.78  & 133.97 & 0.752 \\
    \bottomrule
    \end{tabular}
    }
    \label{tab-add}
\end{table}

Existing synthetic TBPR datasets, usually generated using real data, are used to pre-train TBPR models, serving as complementary augmentation rather than standalone training sources.
For a fair comparison under this conventional setting, we similarly use synthetic data for pre-training and select SynTBPR-S3, which is also grounded in real data\footnote{Notably, these open-source synthetic TBPR datasets~\cite{yang2023towards,shao2023unified,tan2024harnessing,jiang2025modeling} are built from millions of real images; while SynTBPR-S3 employs only ten thousand real images to synthesize images.}.
As shown in Table~\ref{tab-1}, SynTBPR-S3 consistently achieves decent results when integrated into RaSa~\cite{bai2023rasa}, IRRA~\cite{jiang2023cross}, AUL~\cite{li2024adaptive}, despite other datasets being derived from large-scale real datasets. This validates the effectiveness of our pipeline.

Beyond comparisons with methods with synthesizing data, our SynTBPR-S3 pre-trained on RaSa~\cite{bai2023rasa} and AUL~\cite{li2024adaptive} offer decent performance advantages over traditional methods without synthesizing data.
Notably, SynTBPR-S3 is orthogonal to traditional methods—any existing retrieval method can serve as a baseline and further improve through our synthesized data.

\subsection{Ablation studies}

To accelerate the ablation studies, we perform the following experiments on \textbf{a small scale} ($0.03$ million data) within \textbf{the first scenario} on CUHK-PEDES.

\begin{figure}
\centerline{\includegraphics[width=1\linewidth,height=0.38\linewidth]{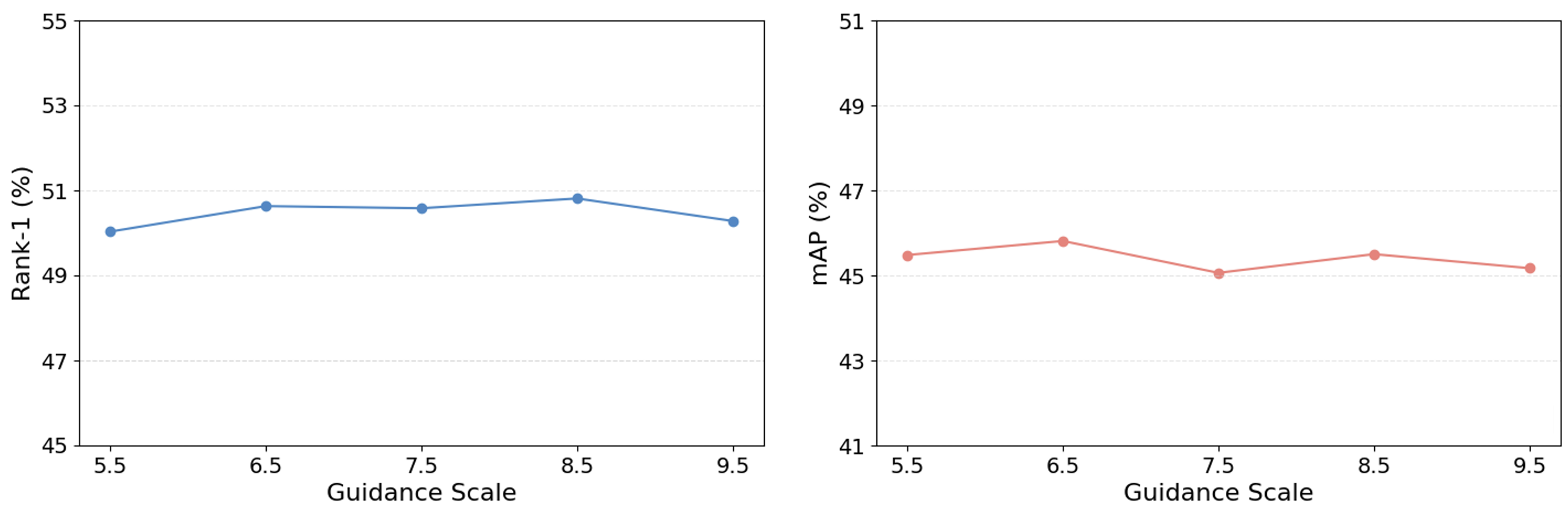}}
\caption{Performance trend under different value of the guidance scale on CUHK-PEDES.}
\label{fig-add}
\end{figure}

\textbf{Effect factors of generating image.}
We explore the inter-class image generation and intra-class image augmentation modules for person image generation.
Three factors usually influence generation process~\cite{fan2024scaling}: generation method, input prompt, and guidance scale.
In addition to the classical Rank-k and mAP metrics, we introduce two additional evaluation criteria to assess the quality of the synthetic images. Specifically, the Fréchet Inception Distance (FID)~\cite{heusel2017gans} measures the similarity between synthetic and real images by computing their feature distribution distance, while the Diversity metric evaluates the variety of the synthetic images.
\textbf{(1) Generation method.}
We use SD~\cite{rombach2022high} for inter-class image generation, referencing \cite{hertz2022prompt} for local/global editing and \cite{cao2023masactrl} for non-rigid editing. Additionally, we replace them with Pixart-$\alpha$ ~\cite{chen2023pixart} for generation and \cite{brooks2023instructpix2pix} for local/global editing. As shown in Table~\ref{tab-add}, different generation methods impact performance. Our adaptive module are expected to enhance TBPR performance with stronger generation methods in the future.
In addition, SD~\cite{rombach2022high} excels in FID, with further editing \cite{hertz2022prompt,cao2023masactrl} optimizing FID\footnote{Notably, the high FID in Table~\ref{tab-add} arises from the presented setting being in the first scenario, where no real data are used, and images are generated using offline SD. Fine-tuning SD on CUHK-PEDES (\emph{i.e.,} our third scenario) improves FID to 47.16.}. Minor Diversity fluctuations may result from limited data scale.
\textbf{(2) Input prompt.}
We develop a plain description template (P. TMPL), and four rough description templates (R. TMPL) focusing on appearance, profession, location and state (No.1-4 respectively) for inter-class image generation. 
The results are shown in Table~\ref{tab-2}.
It can be seen that:
a) The rough description templates (M2) outperform the basic description template (M1), due to higher-quality synthetic images from M2.
b) Different rough templates (M3$\sim$M6) yield consistent results across metrics.
c) Combining both templates (M7) achieves the best results, benefiting from diverse and abundant prompts.
\textbf{(3) Guidance scale.}
Figure~\ref{fig-add} shows performance trends across different guidance scales, revealing fluctuations. Empirical evaluation identifies $8.5$ as the optimal guidance scale for image generation.

\begin{table}
\setlength{\abovecaptionskip}{0.05cm}
    \centering
    \caption{Ablation results of adopting different prompts in inter-class image generation on CUHK-PEDES. FID and Diversity (Div.) are two metrics for evaluating the quality of synthetic images.}
    \scalebox{0.78}{
    \begin{tabular}{l|c|cccc|cccc}
    \toprule
    \rowcolor{gray!25}
    \multirow{2}{*}{}&  & \multicolumn{4}{c|}{R. TMPL} &  &  &  & \\
    \cline{3-6}
    \rowcolor{gray!25}
    & \multirow{-2}{*}{P. TMPL} & No.1 & No.2 & No.3 & No.4 & \multirow{-2}{*}{Rank-1} & \multirow{-2}{*}{mAP} & \multirow{-2}{*}{FID($\downarrow$)} & \multirow{-2}{*}{Div.($\uparrow$)} \\
    \midrule
    M1 & \checkmark & $\times$ & $\times$ & $\times$ & $\times$ &48.00    &42.79    &157.15    & 0.665    \\
    M2 & $\times$ & \checkmark & \checkmark & \checkmark & \checkmark &50.10    &45.04    & 137.59   & 0.739  \\
    \hline
    M3 & $\times$ & \checkmark & $\times$ & $\times$ & $\times$ &50.00    & 44.80   &155.59    & 0.666    \\
    M4 & $\times$ & $\times$ & \checkmark & $\times$ & $\times$ &46.69    &42.19    &144.53    & \textbf{0.784}    \\
    M5 & $\times$ & $\times$ & $\times$ & \checkmark & $\times$ &49.66    &44.82    &140.26    & 0.754   \\
    M6 & $\times$ & $\times$ & $\times$ & $\times$ & \checkmark & 48.98   &43.84    &139.90    &0.747    \\
    \hline
    M7 & \checkmark & \checkmark & \checkmark & \checkmark & \checkmark & \textbf{50.81}   & \textbf{45.50}   & \textbf{137.00}  & 0.751    \\
    \bottomrule
    \end{tabular}
    }
    \label{tab-2}
\end{table}

\begin{figure*}[t]
\setlength{\abovecaptionskip}{0.05cm}
\centerline{\includegraphics[width=1\linewidth,height=0.31\linewidth]{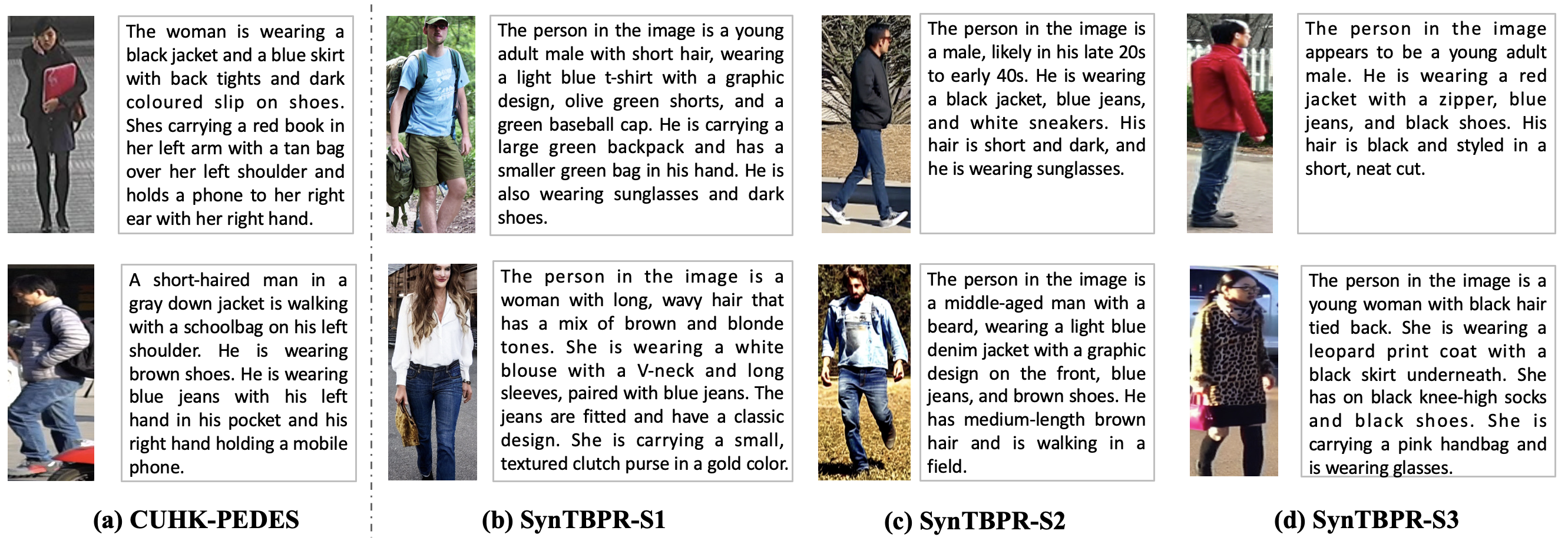}}
\caption{Illustration of real data (a) and synthetic data by our proposed pipeline (b)$\sim$(c). We show example cases from CUHK-PEDES~\cite{li2017person} for real data. The synthetic data labeled as SynTBPR-S1, SynTBPR-S2, and SynTBPR-S3 represent data generated under three training scenarios in our proposed pipeline: S1 (No Data), S2 (Limited Data), and S3 (Abundant Data). For clarity, we display the cropped synthetic images.}
\label{fig-vis}
\end{figure*}

\textbf{Editorial object in intra-class image augmentation.}
Based on the three editing mechanisms, we conduct background, weather, style, and posture edits on synthetic images. 
The results are shown in Table~\ref{tab-3}.
(1) Apart from the augmentation by posture editing, others edits consistently improve performance. 
Posture editing results in synthetic images of generally lower quality and affects performance.
(2) Background editing yields the most significant gains. The resulting data aligns with the natural distribution during inference, where person images have diverse backgrounds. Training TBPR model with these intra-class images enhances background-robust features, improving TBPR.
(3) Weather and style edits also enhance performance.
Although augmented intra-class images from these edits may not perfectly match the natural data distribution during inference, they act as regularization, reducing model overfitting and benefiting TBPR.

\begin{table}
\setlength{\abovecaptionskip}{0.05cm}
    \centering
    \caption{Ablation results of different editorial objects in intra-class image augmentation on CUHK-PEDES.}
    \scalebox{0.92}{
    \begin{tabular}{l|cccc}
    \toprule
    \rowcolor{gray!25}
    Method & Rank-1 & Rank-5 & Rank-10 & mAP \\
    \midrule
    Baseline + Ours (gen.) & 50.81  & 70.31   &78.25   & 45.50 \\
    \hline
    + Ours (aug. w/ background) & 53.90  &\textbf{73.34}  & \textbf{80.43} & 48.64  \\
    + Ours (aug. w/ weather) & 52.65  &  71.56  & 78.91 & 47.56 \\
    + Ours (aug. w/ style) & 52.81 & 71.05   & 79.21  & 47.70  \\
    + Ours (aug. w/ posture) & 51.02 & 70.86   & 78.01 & 45.97 \\
    + Ours (aug.) &\textbf{54.68} & 72.89   &79.99    &\textbf{49.30}   \\
    \bottomrule
    \end{tabular}
    }
    \label{tab-3}
\end{table}

\textbf{Different ways for generating text.}
We explore two ways of text generation: constant instruction guidance and variable instruction guidance.
The generated texts are denoted as $T_c$ and $T_v$, respectively.
Based on the results in Table~\ref{tab-4}, the following observations are noted.
(1) Among the two ways of text generation (M1$\sim$M4), the best outcomes are obtained with variable instruction guidance on InternVL~\cite{chen2024internvl} (M2). The variable instruction guidance, employed in the powerful MLLM (InternVL~\cite{chen2024internvl}), produces higher-quality text with diverse structures over constant instruction guidance.
(2) Furthermore, pairing all generated texts with corresponding images (M5) leads to optimal results, emphasizing their complementarity.


\begin{table}
\setlength{\abovecaptionskip}{0.05cm}
    \centering
    \caption{Ablation results for different text generation ways. CLIP-score($\uparrow$) measures synthetic text quality on CUHK-PEDES.}
    \scalebox{0.78}{
    \begin{tabular}{l|c|ccc|ccc}
    \toprule
    \rowcolor{gray!25}
    \multirow{2}{*}{} & $T_{c}$ & $T_{v}^{InternVL}$ & $T_{v}^{MiniCPM}$ & $T_{v}^{QwenVL}$ & Rank-1 & mAP & CLIP-score\\
    \midrule
    M1 & \checkmark & $\times$ & $\times$ & $\times$ & 40.46   & 35.23  & 27.282  \\
    M2 & $\times$ & \checkmark & $\times$ & $\times$ & 42.38   &37.86   & \textbf{31.968}  \\
    M3 & $\times$ & $\times$ & \checkmark & $\times$ & 40.37   & 36.33  &31.553   \\
    M4 & $\times$ & $\times$ & $\times$ & \checkmark &38.27   & 34.65  &31.367   \\
    M5 & \checkmark & \checkmark & \checkmark & \checkmark &\textbf{50.81}   &\textbf{45.50}  &  30.542 \\
    \bottomrule
    \end{tabular}
    }
    \label{tab-4}
\end{table}

\textbf{Noise analysis.}
Synthetic data contains noise, necessitating noise analysis (Table~\ref{tab-5}). Denoising during generation improves performance (M1$\sim$M4), with significant gains in image denoising (M2) but only slight improvements in text denoising (M3). Since texts are generated after person images, image noise has a greater impact than text noise. Additionally, in noise-robust learning for retrieval training (M5$\sim$M8), only label smoothing (M6) proves effective.

\begin{table}[t]
\setlength{\abovecaptionskip}{0.05cm}
    \centering
    \caption{Ablation results of noise analysis on CUHK-PEDES.}
    \scalebox{0.92}{
    \begin{tabular}{l|cc|ccc|cc}
    \toprule
    \rowcolor{gray!25}
    \multirow{2}{*}{} & \multicolumn{2}{c|}{Generation} & \multicolumn{3}{c|}{Retrieval} &  &  \\
    \rowcolor{gray!25}
    & Img D. & Txt D. & Label-S & GMM & C-Mask & \multirow{-2}{*}{Rank-1} & \multirow{-2}{*}{mAP} \\
    \midrule
    M1 & $\times$ & $\times$ & \checkmark & $\times$ & $\times$ &43.29   &38.51   \\
    M2 & \checkmark & $\times$ & \checkmark & $\times$ & $\times$  & \textbf{50.81}   & \textbf{45.50} \\
    M3 & $\times$ & \checkmark & \checkmark & $\times$ & $\times$ &43.73   &38.71   \\
    M4 & \checkmark & \checkmark & \checkmark & $\times$ & $\times$ & 50.68  & 45.33  \\
    \hline
    M5 & \checkmark & $\times$ & $\times$ & $\times$ & $\times$ & 45.89  &41.73   \\
    M6 & \checkmark & $\times$ & \checkmark & $\times$ & $\times$ & \textbf{50.81}   & \textbf{45.50}   \\
    M7 & \checkmark & $\times$ & $\times$ & \checkmark & $\times$ &45.44   & 41.29   \\
    M8 & \checkmark & $\times$ & $\times$ & $\times$ & \checkmark & 45.74  &41.67   \\
    \bottomrule
    \end{tabular}
    }
    \label{tab-5}
\end{table}

\subsection{Visualization results}

We present visualization results of synthetic data by our proposed pipeline and real data in Fig~\ref{fig-vis}.
More visualization can be seen in the supplementary material.

%% file: sec/5_limitation.tex
\section{Further discussion}


\subsection{More application exploration}

We have conducted comprehensive studies on synthetic data across three scenarios.
Additionally, we explore further potential applications of synthetic data.

\begin{table}
\setlength{\abovecaptionskip}{0.1cm}
    \centering
    \caption{Results of cross-domain TBPR (\emph{i.e.}, source: CUHK-PEDES $ \rightarrow $ target: ICFG-PEDES).}
    \scalebox{0.95}{
    \begin{tabular}{l|c|ccc}
    \toprule
    \rowcolor{gray!25}
    Method & \# target data & Rank-1 & Rank-5 & Rank-10 \\
    \midrule
    APTM (MM23)~\cite{yang2023towards} & \multirow{3}{*}{0}  &  46.20 &  65.13 &  72.59 \\
    RaSa (IJCAI23)~\cite{bai2023rasa} &  & 50.59 & 67.46 &  74.09 \\
    AUL (AAAI24)~\cite{li2024adaptive} &  &  49.29 &  67.46 &  74.42 \\
    \hline
    Baseline &   & 46.19   &64.49   & 71.94  \\
     + Ours & \multirow{-2}{*}{8} &  50.80     & 67.98       & 74.84     \\
    \bottomrule
    \end{tabular}
    }
    \label{SM-tab8}
\end{table}

\begin{table}[t]
\setlength{\abovecaptionskip}{0.1cm}
    \centering
    \caption{Results of TBPR in rare environment.}
    \scalebox{0.95}{
    \begin{tabular}{l|cccc}
    \toprule
    \rowcolor{gray!25}
    Method  & Rank-1 & Rank-5 & Rank-10 & mAP \\
    \midrule
    IRRA(CVPR23)~\cite{jiang2023cross} & 68.67 & 87.63   & 91.95   & 63.94   \\
    AUL(AAAI24)~\cite{li2024adaptive} & 70.35  & 87.10    & 92.24    & 62.23    \\
    \hline
    Baseline   & 70.92 & 88.21   & 92.73  & 65.80  \\
    + Ours   &72.80 & 89.20    &92.97   & 68.28  \\
    \bottomrule
    \end{tabular}
    }
    \label{SM-tab9}
\end{table}

\begin{table}
\setlength{\abovecaptionskip}{0.1cm}
    \centering
    \caption{Result of computational complexity. Latency refers to the time (measured in seconds) required to generate an image. These results are tested on an H800 GPU.}
    \scalebox{0.8}{
    \begin{tabular}{l|cc|cc}
    \toprule
    \rowcolor{gray!25}
     & \multicolumn{2}{c|}{Image generation} & \multicolumn{2}{c}{Text generation} \\
    \cline{2-5}
    \rowcolor{gray!25}
    \multirow{-2}{*}{Method} & Latency(s) & \# Model Param(B) & Words Per Second & \# Model Param(B) \\
    \midrule
    Ours &0.96  &1.06   &45.85  & 25.5 \\
    \bottomrule
    \end{tabular}
    }
    \label{tab-overhead}
\end{table}

\textbf{Cross-domain TBPR.}
We successfully explore the technical feasibility of achieving TBPR with limited real data (\emph{e.g.}, $8$ image-text pairs in scenario 2).
Taking a step further, it could be applied in cross-domain TBPR in which TBPR model trained in source data are directly applied in target data.

Cross-domain TBPR holds significant practical value. 
For instance, there is a powerful TBPR model trained on person data from one environment (called source data).
However, there arises a need to conduct TBPR in a new environment where person data (called target data) is typically scarce due to the high acquisition costs.
In such scenario, a TBPR model with robust cross-domain capability becomes essential.
Our proposed pipeline empower the TBPR model with decent cross-domain ability. 
Through our pipeline, abundant synthetic target data can be obtained on the basis of just $8$ real target image-text pairs, which are cost-effective and practical to collect.
The results in Table \ref{SM-tab8} show the effectiveness of our proposed pipeline when integrated into baseline.
Importantly, our pipeline are adaptable to various TBPR models, endowing them with cross-domain capability as well.

\textbf{TBPR in rare environment.}
There is a potential application for TBPR.
At times, we need to find person captured in rare environments, such as high winds or heavy snow.
However, gathering a large number of person data in these rare environments for model training is challenging due to their scarcity.
A TBPR model trained without an adequate amount of person data in these rare environments tends to exhibit inferior performance when applied in such environments.
One potential solution is to generate additional person data in rare environments to supplement the training data.
We conduct experiments to evaluate the practicability of our proposed pipelines for TBPR in rare environments.

Specifically, we take the snow environment as an example. 
The test data containing person images in snowy environment is lacking. 
Consequently, we generate such data based on the CUHK-PEDES test dataset and then manually select high-quality data to augment the CUHK-PEDES test data, resulting in a new test dataset termed \emph{CUHK-PEDES (snow) test data}. 
This new test dataset can simulate real rare environments, based on which we conduct experiments involving three models, including the baseline model integrated with our proposed pipelines, IRRA~\cite{jiang2023cross} and AUL~\cite{li2024adaptive}. 
As shown in Table~\ref{SM-tab9}, the baseline model, when coupled with our proposed pipelines, achieve the best results. 
Our proposed pipelines introduce synthetic person data in snowy environments as training data, bridging the gap between training and test data, thus enhancing performance.
In addition, our proposed pipelines is actually a data generation paradigm that can be implemented in any TBPR model to enhance performance in rare environments.

\subsection{Limitation discussion}
\label{LD}

As a key aspect of empirically validating synthetic data for TBPR, the limitations of generating such data are discussed here.
Specifically, there are two limitations.
\textbf{(1) Domain discrepancy}. 
Despite advances in generative models, they are typically pre-trained on web-crawled data containing diverse objects. As a result, these models generate TBPR data with a distribution gap from real data, which impacts retrieval performance.
\textbf{(2) Computational overhead}.
It is widely known that generative models have high computational costs, making them unsuitable for resource-limited scenarios.
We show the computational complexity in Table~\ref{tab-overhead}.

Although the above limitations exist, several points need to be highlighted.
(1) For domain discrepancy, with the rapid advancement of generation technologies, the gap between generated and real data has significantly decreased compared to earlier methods (manually simulating virtual environments and 3D persons~\cite{sun2019dissecting,wang2020surpassing}, or using GANs~\cite{zheng2019joint}).
On the other hand, \textbf{we have already made initial attempts to address domain discrepancy.} These efforts include designing a prompt construction strategy to generate diverse and large-scale data, and exploring noise-robust learning for TBPR models. Unlike other methods that rely on real textual descriptions to guide image generation~\cite{yang2023towards}, our method aims to be more efficient and scalable.
As shown in Fig.~\ref{fig-vis}, our generated data is nearly equivalent in quality to real data. 
Moving forward, we consider to develop generative models specifically tailored for TBPR to further enhance performance.
(2) For computational overhead, while generating data introduces computational costs, it remains a necessary and optimal solution when real data is scarce or unavailable.
\textbf{To further mitigate the problem, we release our generated data for immediate use, bypassing costly generation.}
Moreover, research on lightweight generative models will be considered.

%% file: sec/6_conclusion.tex
\section{Conclusion}

This work aims not to develop a new model, rather, we conduct an empirical study to validate synthetic data from generative AI models for TBPR.
We aim to address critical issues of previous methods: labor-intensive, privacy-sensitive, diversity-deficient, and exploration-restricted.
For this, we develop an inter-class image generation pipeline and an intra-class image augmentation pipeline.
They serve as the foundation for a comprehensive study of synthetic data effectiveness across three representative scenarios, along with an initial exploration of more realistic scenarios.
Also, considering the inherent noise in synthetic data, we experimentally explore noise-robust learning strategies to enhance TBPR performance.

%% file: sec/7_supplementary.tex
\section{Methodology}
\label{method}

\subsection{Descriptor lists}
\label{DL}
We design both a plain description template and four rough description templates for inter-class image generation. 
By replacing $\{*\}$ in these templates with elements from the descriptor lists, we obtain the prompt to steer text-driven image synthesis models towards generating images.
The descriptor lists are obtained by instructing ChatGPT to select and augment the tags sourced from~\cite{huang2023tag2text,yang2023towards}.
The descriptor lists are shown in Fig.~\ref{SM-fig1}.

\subsection{LLM-extend prompts}
\label{LP}

For inter-class image generation, we develop a LLM-extended prompts, for which we instruct Qwen2~\cite{yang2024qwen2} to extend the primary prompts derived from rough description templates into a final set of diverse prompts.
Specifically, the instruction is designed as follows:
\begin{tcolorbox}[colback=white, colframe=black, opacityframe=0.5]
\small \emph{In the style of "The [\{man/woman\}] is []." This is input: \{primary prompt\}, You can remove some clothing or add some accessories. \{sub-instruction\}. Keep the final description under 77 words.}
\end{tcolorbox}

Here, \emph{\{primary prompt\}} is replaced with one of four primary prompts derived from rough description templates.
\emph{\{sub-instruction\}} is then randomly replaced with one of the following instructions:
\begin{tcolorbox}[colback=white, colframe=black, opacityframe=0.5]
\begin{itemize}
\small \item \emph{Expand this description with more details about the person's appearance, clothing, and context.}
\small \item \emph{Provide more details on the person's appearance and clothing, including accessories and context.}
\small \item \emph{Enhance the description by adding details about the setting and the person's appearance and clothing.}
\small \item \emph{Expand this description by including details on the person's actions, emotions, and surroundings.}
\small \item \emph{Add details about the weather and environment to enrich the description of the person's appearance and clothing.}
\small \item \emph{Add details about the person's facial expressions, hairstyle, and any noticeable features.}
\end{itemize}
\end{tcolorbox}

The instructions above are generated by ChatGPT and contribute to enhancing the diversity of the final prompts.

\begin{figure}
\setlength{\abovecaptionskip}{0.1cm}
\centerline{\includegraphics[width=1\linewidth,height=0.65\linewidth]{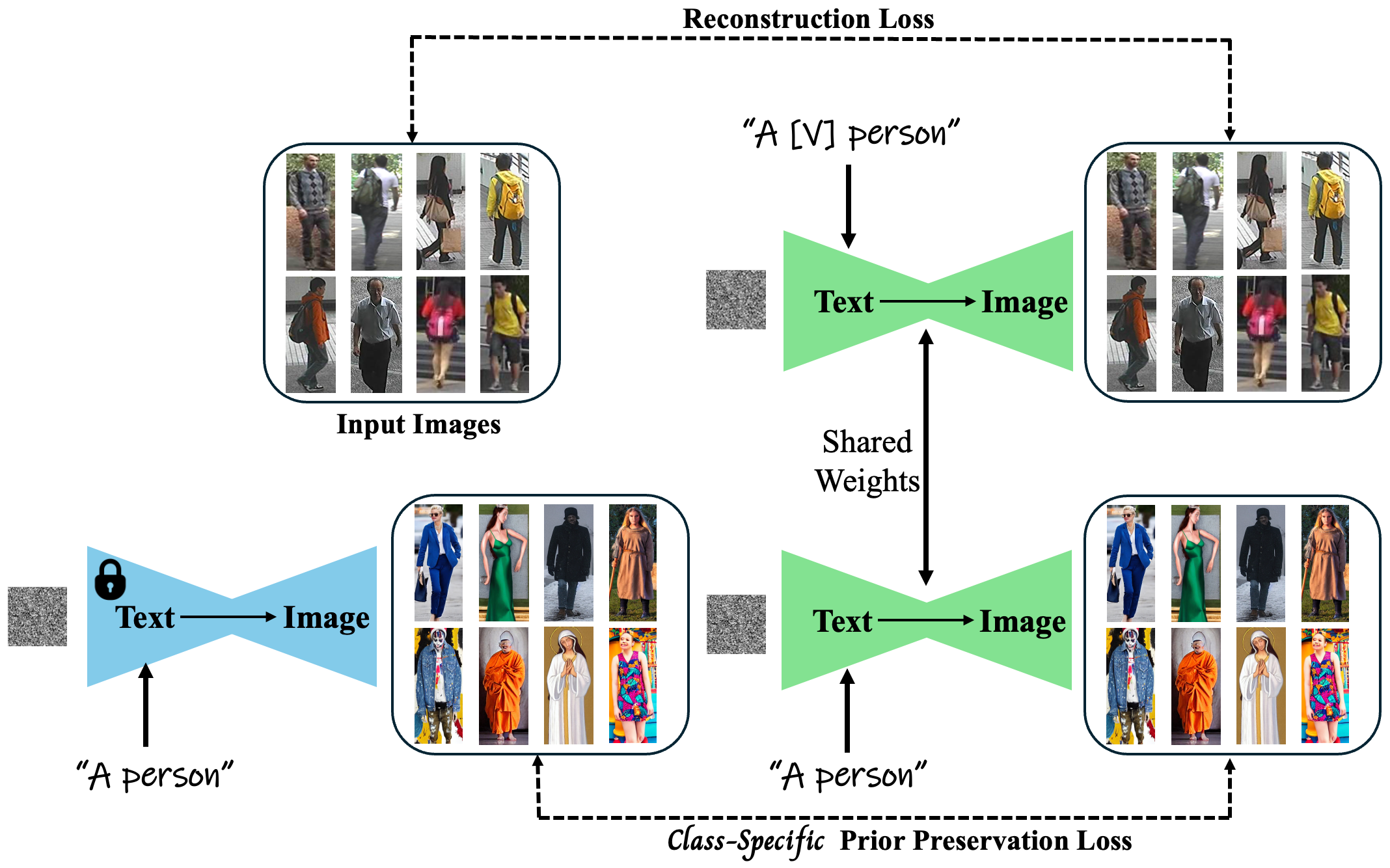}}
\caption{Illustration of stable diffusion fine-tuning using limited person data and prior-preservation loss in Eq. (1) from the main text. This figure corresponds to~\cite{ruiz2023dreambooth}.}
\label{SM-eq1_illu}
\end{figure}

\begin{figure*}
\setlength{\abovecaptionskip}{0.15cm}
\centerline{\includegraphics[width=0.95\linewidth,height=0.85\linewidth]{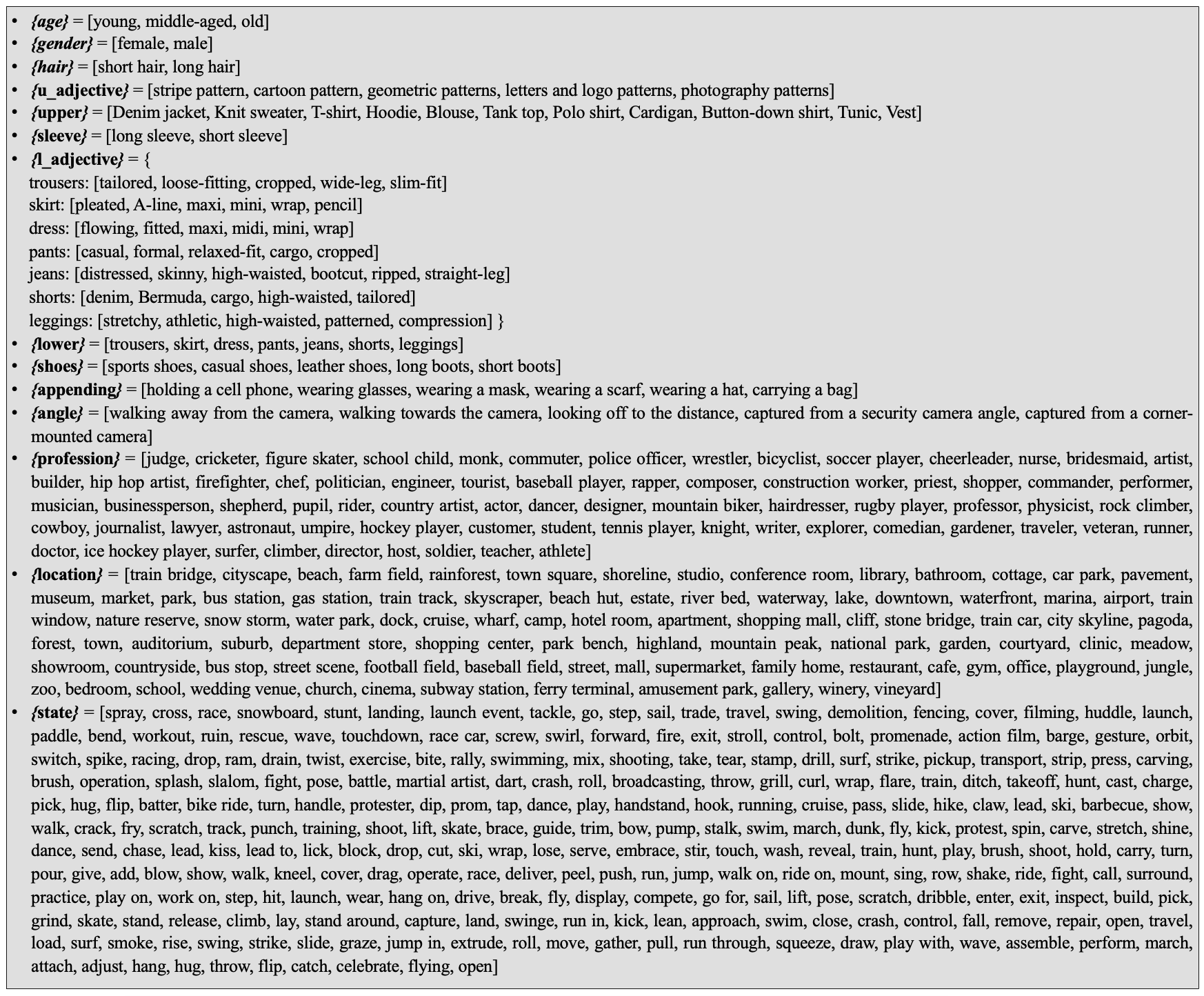}}
\caption{Descriptor lists used in the rough description templates for inter-class image generation.}
\label{SM-fig1}
\end{figure*}

\subsection{Template pool}
\label{Tmp-P}
\begin{figure*}
\setlength{\abovecaptionskip}{0.15cm}
\centerline{\includegraphics[width=1\linewidth,height=0.46\linewidth]{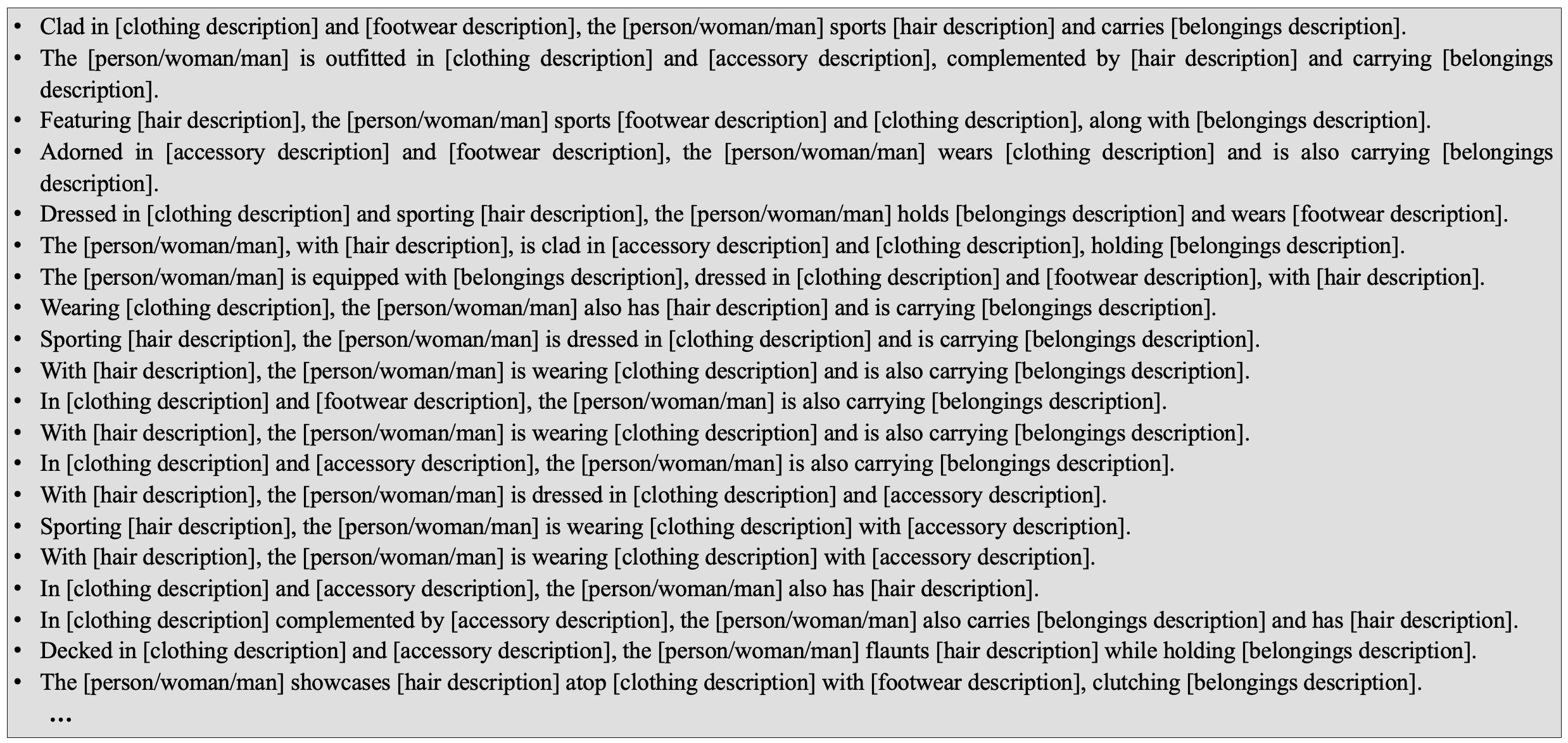}}
\caption{Some templates used in generating the textual description of synthetic image.}
\label{TP}
\end{figure*}

We design a variable instruction to leverage MLLMs to generate the textual description of the synthetic image.
In the variable instruction, we need to replace \emph{\{template\}} from a template pool.
Referring to MLLM4Text-ReID~\cite{tan2024harnessing}, we utilize ChatGPT through multiple rounds of dialogue and iterative optimization and create a collection of $121$ templates.
Fig.~\ref{TP} showcases some templates considering space limitation, while the entire collection is available in the project code.

\section{Experiments}
\label{experiment}

\begin{figure*}
\setlength{\abovecaptionskip}{0.05cm}
\centerline{\includegraphics[width=0.9\linewidth,height=0.33\linewidth]{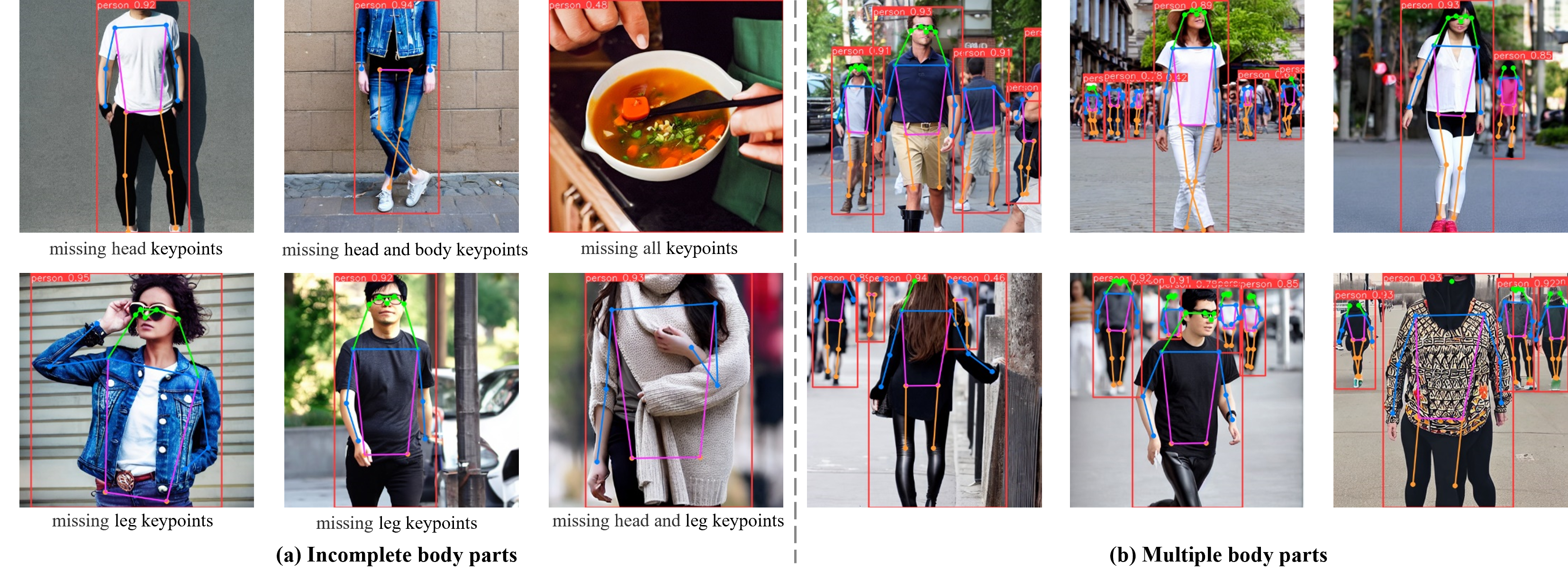}}
\caption{Illustration of noisy images (uncropped). Noise typically arises from incomplete body parts or multiple body parts during generating images.}
\label{SM-syndataI}
\end{figure*}

\subsection{Visualization results}
\label{Vis-R}
\begin{figure*}
\setlength{\abovecaptionskip}{0.1cm}
\centerline{\includegraphics[width=1\linewidth,height=0.33\linewidth]{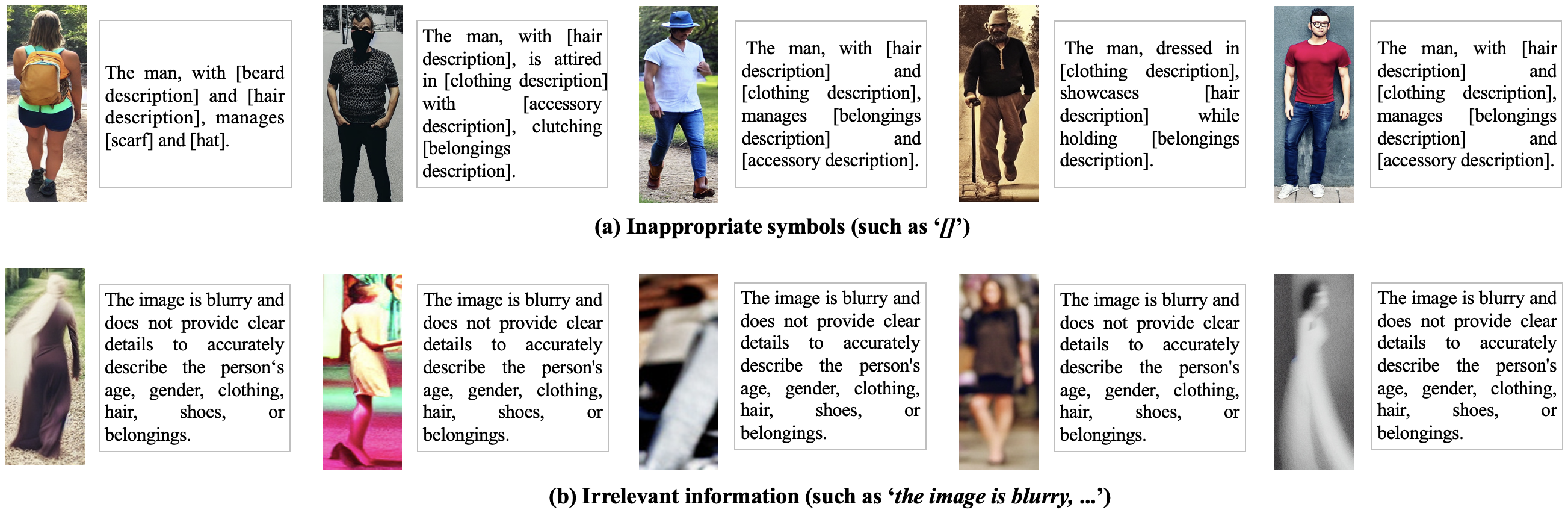}}
\caption{Illustration of noisy texts. Noise usually manifests as inappropriate symbols or irrelevant information within the text.}
\label{SM-syndataT}
\end{figure*}

\begin{figure*}
\setlength{\abovecaptionskip}{0.1cm}
\centerline{\includegraphics[width=1\linewidth,height=0.82\linewidth]{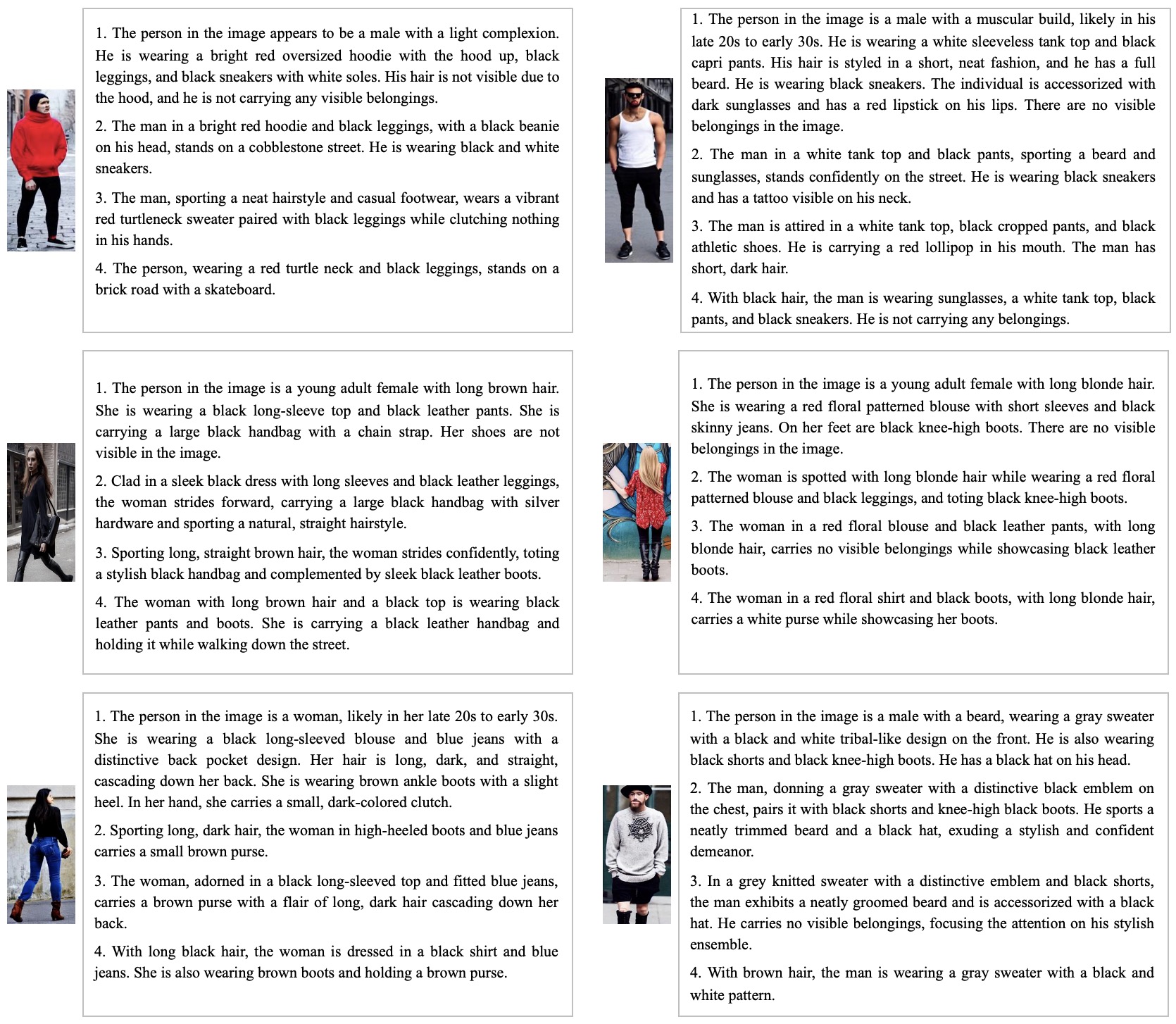}}
\caption{Illustration of synthetic data. The images are generated from the proposed inter-class image generation module under scenario $1$ and the textual descriptors obtained from the text generation pipeline are listed in sequence as \small{$T_c$}, \small{$T_v^{InternVL}$}, \small{$T_v^{MiniCPM}$} and \small{$T_v^{QwenVL}$}.}
\label{SM-syndata}
\end{figure*}

\begin{figure*}
\setlength{\abovecaptionskip}{0.05cm}
\centerline{\includegraphics[width=0.85\linewidth,height=1.3\linewidth]{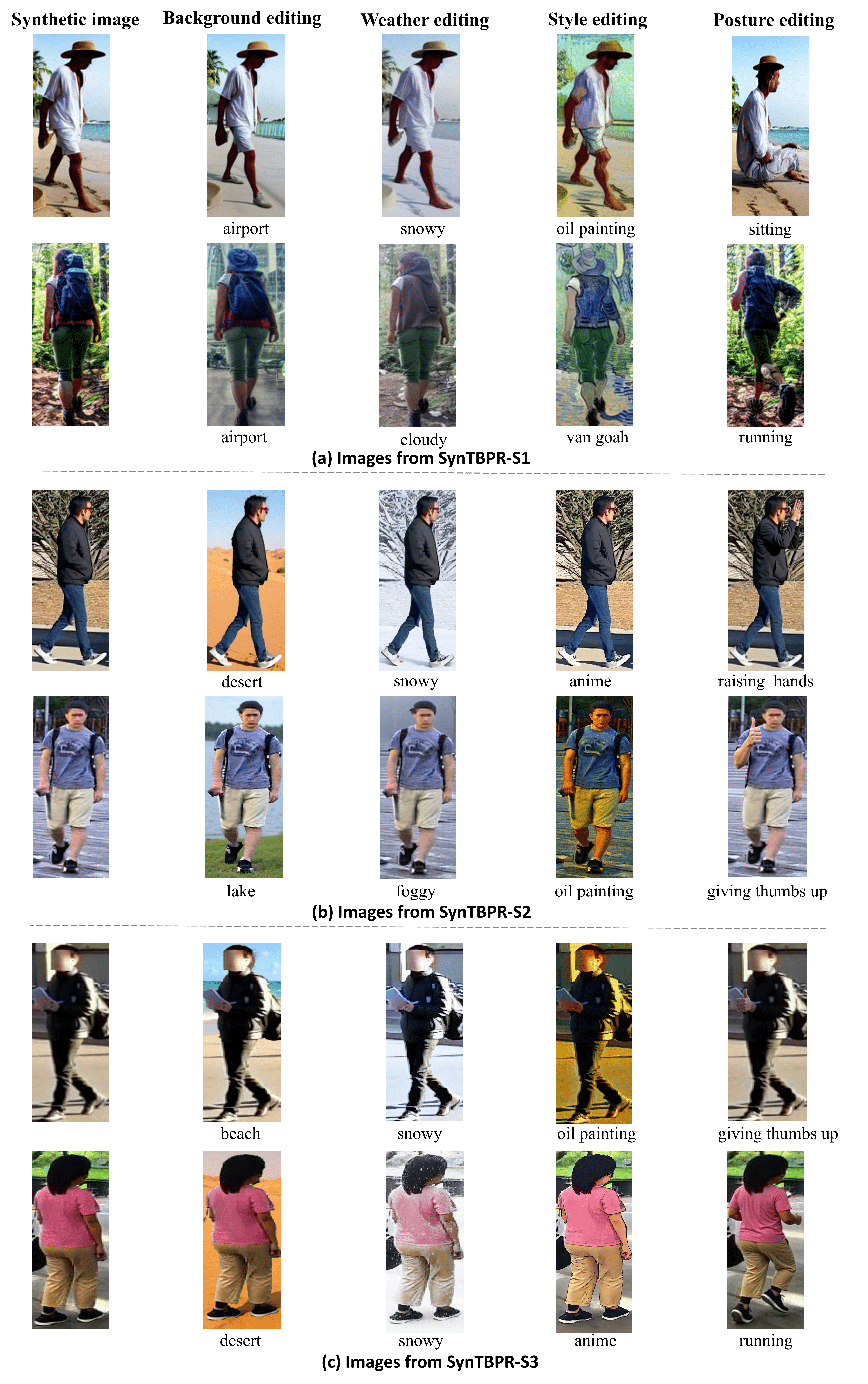}}
\caption{Illustration of editing images. We show the editing images under the first scenario, second scenario, and third scenario, respectively.}
\label{SM-edit}
\end{figure*}

\begin{figure*}
\setlength{\abovecaptionskip}{0.05cm}
\centerline{\includegraphics[width=0.97\linewidth,height=1.2\linewidth]{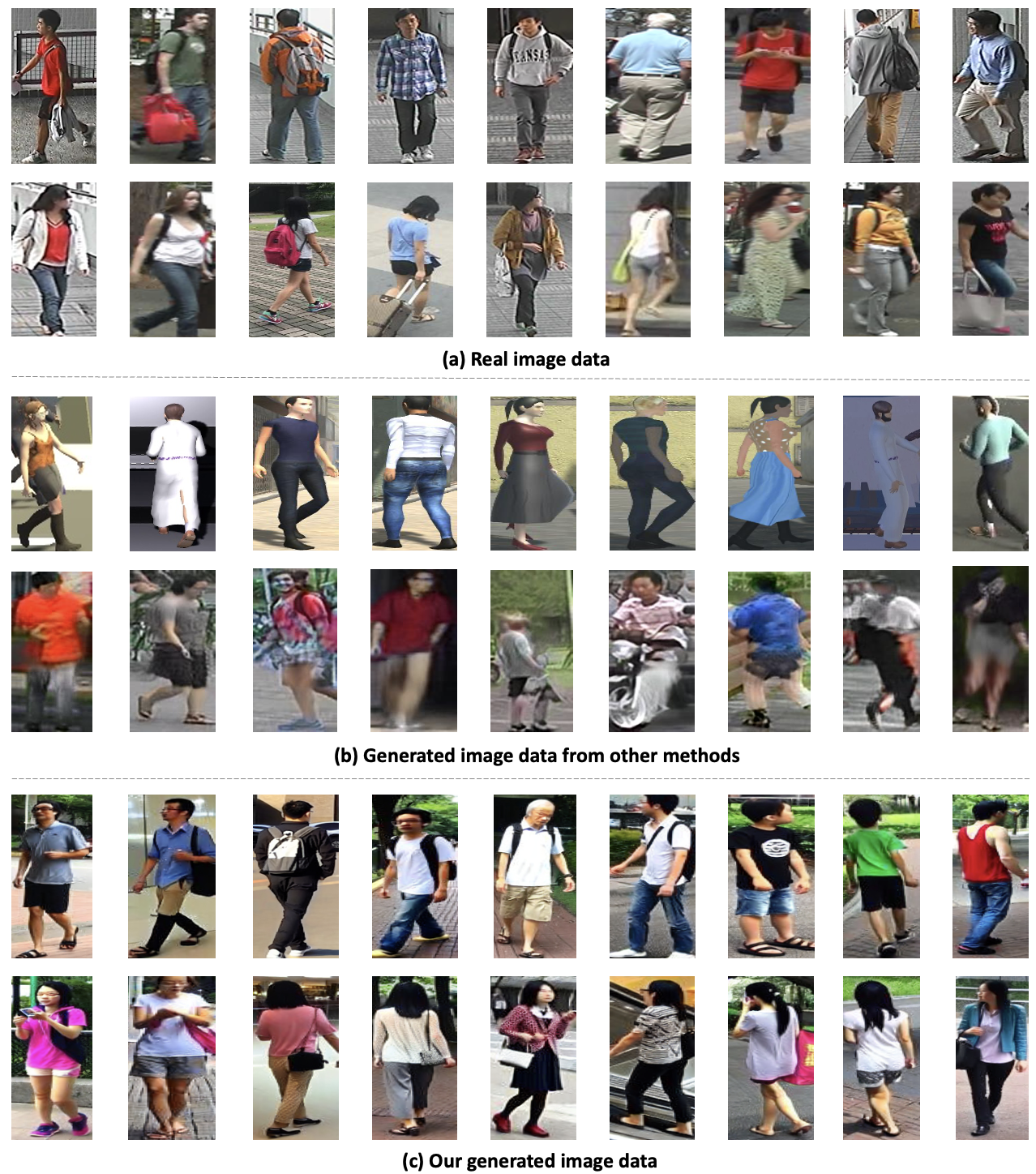}}
\caption{Illustration of real data (CUHK-PEDES) and synthetic data. Synthetic data is shown from other method~\cite{zheng2019joint,sun2019dissecting} and our proposed pipeline.}
\label{SM-allillu}
\end{figure*}

\begin{figure*}
\setlength{\abovecaptionskip}{0.1cm}
\centerline{\includegraphics[width=0.82\linewidth,height=0.5\linewidth]{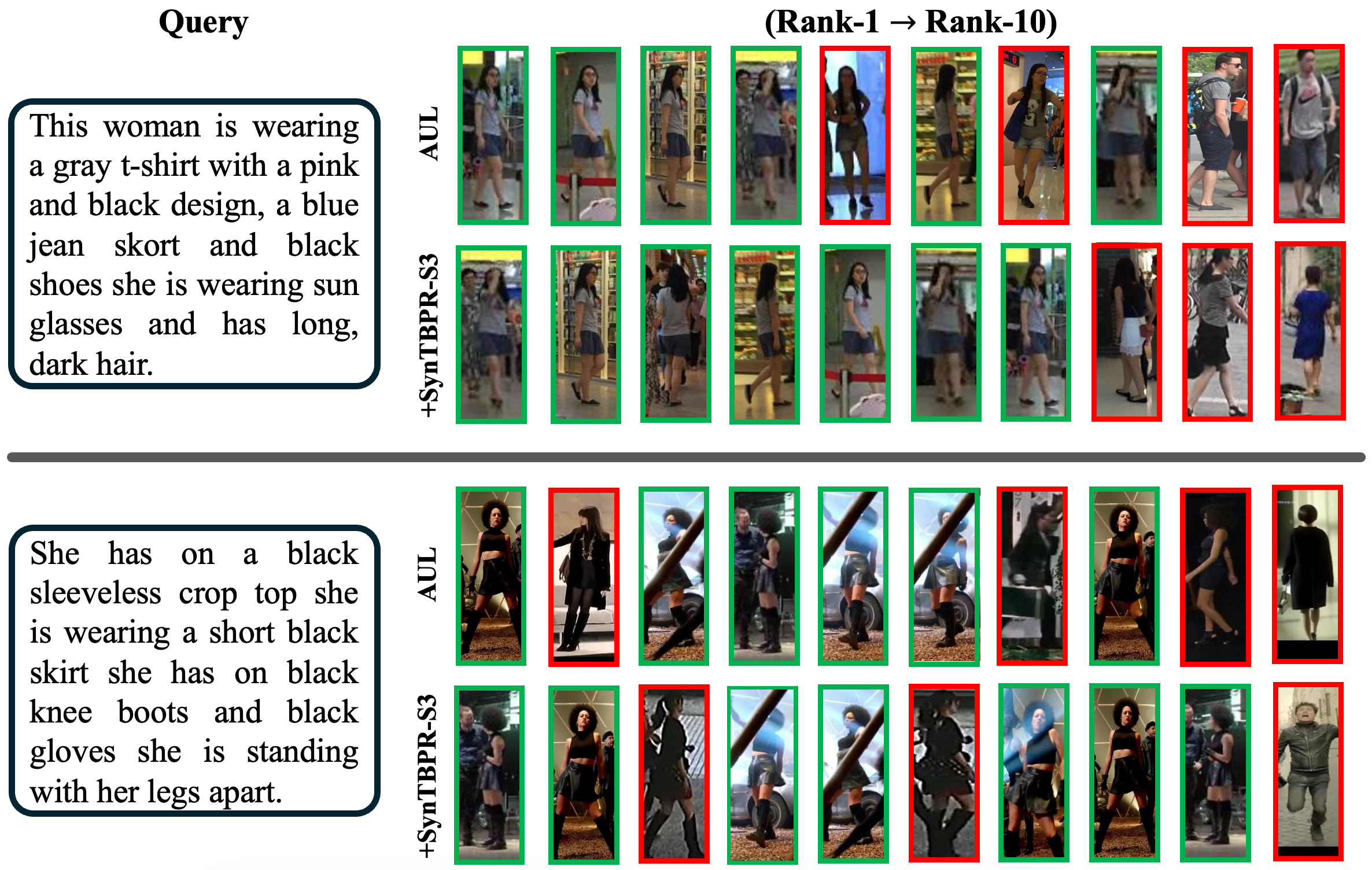}}
\caption{Visualization of retrieval results (R@1 to R@10) in CUHK-PEDES. Each row represents a query-response group. Green and red bounding
boxes indicate correct and incorrect matches, respectively. AUL~\cite{li2024adaptive} is used as the baseline.}
\label{SM-retrievalV}
\end{figure*}

We present visualization results, including an illustration of Eq. (1) from the main text in Fig.~\ref{SM-eq1_illu},
examples of noisy data in Fig.~\ref{SM-syndataI} and Fig.~\ref{SM-syndataT}, editing images in Fig.~\ref{SM-edit}, synthetic data in Fig.~\ref{SM-syndata}, a comparison between real and synthetic images in Fig.~\ref{SM-allillu}, and retrieval visualizations in Fig.~\ref{SM-retrievalV}. 
We display the cropped synthetic images for clarity, unless otherwise specified.

